\pdfoutput=1
\documentclass[11pt]{article}

\usepackage[preprint]{acl}

\usepackage{times}
\usepackage{latexsym}
\usepackage{booktabs} % For \toprule, \midrule,
\usepackage{array}    % For better table formatting
\usepackage{graphicx} % For \resizebox if needed
\usepackage{multirow}    % For cells spanning multiple rows
\usepackage{makecell}    % For line breaks in cells
\usepackage{threeparttable} 
\usepackage{tabularx}
\usepackage{amsmath}

\usepackage[most]{tcolorbox} % 含 skins 等
\usepackage{enumitem}        % 紧凑的 itemize

\usepackage{listings}
\usepackage{xcolor} % 可选，用于高亮

\usepackage{adjustbox}

\usepackage{adjustbox}
\usepackage{subfig}
\usepackage{soul}
\usepackage{colortbl}
\usepackage{geometry}
\usepackage{tcolorbox}
\usepackage{tabularx}
\usepackage{longtable}
\usepackage{array}
\usepackage{booktabs}
\usepackage{amsmath}

\lstdefinelanguage{json}{
    basicstyle=\ttfamily,
    numbers=left,
    numberstyle=\tiny\color{gray},
    stepnumber=1,
    numbersep=5pt,
    showstringspaces=false,
    breaklines=true,
    frame=lines,
    backgroundcolor=\color{white},
    literate=
     *{0}{{{\color{blue}0}}}{1}
      {1}{{{\color{blue}1}}}{1}
      {2}{{{\color{blue}2}}}{1}
      {3}{{{\color{blue}3}}}{1}
      {4}{{{\color{blue}4}}}{1}
      {5}{{{\color{blue}5}}}{1}
      {6}{{{\color{blue}6}}}{1}
      {7}{{{\color{blue}7}}}{1}
      {8}{{{\color{blue}8}}}{1}
      {9}{{{\color{blue}9}}}{1}
      {:}{{{\color{black}:}}}{1}
      {,}{{{\color{black},}}}{1}
      {"}{{{\color{red}"}}}{1}
      {[}{{{\color{black}[}}}{1}
      {]}{{{\color{black}]}}}{1}
      {\{}{{{\color{black}\{}}}{1}
      {\}}{{{\color{black}\}}}}{1},
}

\usepackage[T1]{fontenc}
\usepackage[utf8]{inputenc}

\usepackage{microtype}

\usepackage{inconsolata}

\usepackage{graphicx}
\newcommand\blfootnote[1]{%
\begingroup
\renewcommand\thefootnote{}\footnote{#1}%
\addtocounter{footnote}{-1}%
\endgroup
}

\tcbuselibrary{skins,breakable}
\usepackage{fvextra} % Verbatim with breaklines + commandchars

\definecolor{DoctorColor}{HTML}{2E7D32}  % green
\definecolor{PatientColor}{HTML}{7B1FA2} % purple

\newenvironment{convverbatim}{%
  \Verbatim[
    breaklines=true,
    breakanywhere=true,
    commandchars=\\\{\}, % allow \textcolor{...}{...}
    fontsize=\tiny
  ]%
}{\endVerbatim}

\newcommand{\NSTAGE}{%
  \noindent\smash{\rule{\linewidth}{0.4pt}}% 线条本身不占高度
  \strut% 占用标准行高，确保与普通文本行一致
}

\title{DischargeSim: A Simulation Benchmark for Educational

Doctor–Patient Communication at Discharge}

\author{%
  Zonghai Yao~\thanks{indicates equal contribution}$^{1,2}$, 
  Michael Sun~\footnotemark[1]$^{2}$,
  Won Seok Jang$^{1,3}$,  
  Sunjae Kwon$^{1,2}$, 
  Soie Kwon$^{4}$, 
  Hong Yu$^{1,2,3}$
  \\
  $^{1}$Center for Healthcare Organization and Implementation Research, VA Bedford Health Care  \\
  $^{2}$Manning College of Information and Computer Sciences, UMass Amherst, MA, USA \\
  $^{3}$Miner School of Computer and Information Sciences, UMass Lowell, MA, USA \\
  $^{4}$Department of Internal Medicine, Chung-Ang University, Seoul, Republic of Korea \\
  \texttt\ \{\href{mailto:zonghaiyao@umass.edu}{zonghaiyao}, \href{mailto:mssun@umass.edu}{mssun}\}@{umass.edu}
  \\
}

\begin{document}
\maketitle
\begin{abstract}

Discharge communication is a critical yet underexplored component of patient care, where the goal shifts from diagnosis to education. While recent large language model (LLM) benchmarks emphasize in-visit diagnostic reasoning, they fail to evaluate models’ ability to support patients after the visit. We introduce DischargeSim, a novel benchmark that evaluates LLMs on their ability to act as personalized discharge educators. DischargeSim simulates post-visit, multi-turn conversations between LLM-driven DoctorAgents and PatientAgents with diverse psychosocial profiles (e.g., health literacy, education, emotion). Interactions are structured across six clinically grounded discharge topics and assessed along three axes: (1) dialogue quality via automatic and LLM-as-judge evaluation, (2) personalized document generation including free-text summaries and structured AHRQ checklists, and (3) patient comprehension through a downstream multiple-choice exam. 
Experiments across 18 LLMs reveal significant gaps in discharge education capability, with performance varying widely across patient profiles. 
Notably, model size does not always yield better education outcomes, highlighting trade-offs in strategy use and content prioritization. 
DischargeSim offers a first step toward benchmarking LLMs in post-visit clinical education and promoting equitable, personalized patient support.~\footnote{The source code is released at: \url{https://github.com/michaels6060/DischargeSim} with CC-BY-NC 4.0 license.}.

\end{abstract}

\begin{figure*}[t]
\centering
\includegraphics[scale=0.63]{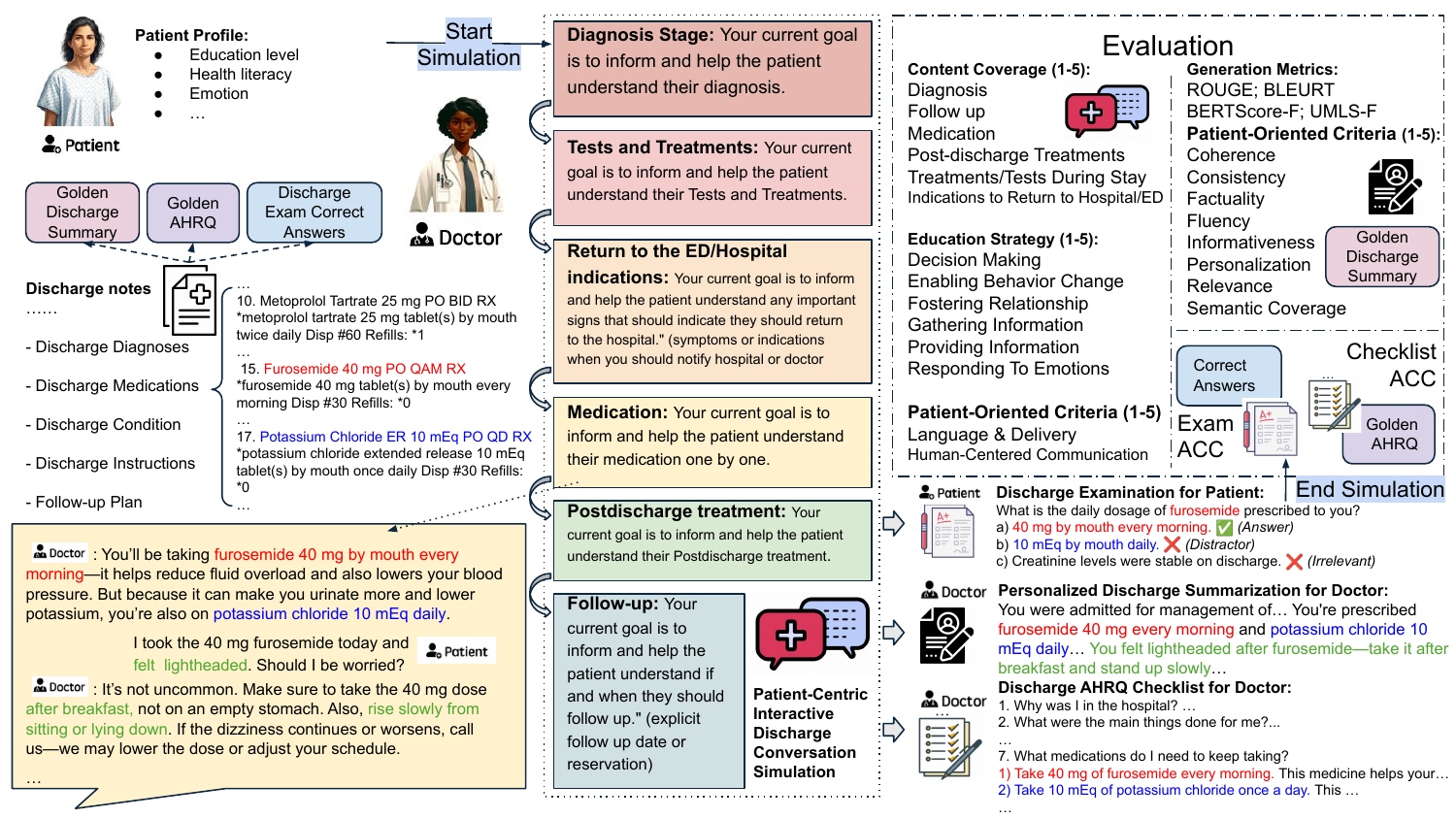} \hfill
\caption{\textbf{DischargeSim Overview.} This figure illustrates the full workflow of DischargeSim, a simulation framework for evaluating post-visit doctor-patient discharge conversations. The simulation begins with a \textit{PatientAgent} defined by its education level, health literacy, and emotional state. 
The conversation progresses through six content stages: Diagnosis, Tests and Treatments, Return-to-Hospital Indicators, Medication, Post-discharge Treatment, and Follow-up, each with stage-specific education goals. 
The central panel illustrates example turns where the DoctorAgent employs communication strategies such as simplification and emotional support.
The right panel depicts several evaluation components that enable comprehensive simulation and evaluation of LLM-based discharge communication.
}
\label{fig:overview}
\end{figure*}

\section{Introduction}

\blfootnote{$\dagger$ To appear in proceedings of the Main Conference on Empirical Methods in Natural Language Processing (EMNLP) 2025}

Discharge communication plays a crucial role in ensuring patient safety and promoting long-term recovery. Upon leaving the hospital, patients must understand a range of instructions, including medications, follow-up appointments, return precautions, and post-discharge care. Yet studies show that 40\% to 80\% of patients forget or misunderstand this information shortly after discharge~\cite{kessels2003patients, richard2017communication, federman2018challenges}. Miscommunication at this stage contributes to poor adherence, increased readmissions, and preventable complications~\cite{zhao2018barriers, weerahandi2018predictors}.

While recent clinical NLP benchmarks, such as AgentClinic~\cite{agentclinic2024}, AMIE~\cite{tu2025towards}, and HealthBench~\cite{arora2025healthbench}, focus on diagnostic reasoning during visits, they largely overlook the discharge phase, a stage that prioritizes education over inference. Some recent work~\cite{paniniqa2023} tried to address discharge understanding but is limited to single-turn QA, lacking dialogue structure, patient diversity, and downstream evaluation.

To fill this gap, we introduce DischargeSim, a comprehensive benchmark that evaluates large language models (LLMs) in the role of discharge educators. As shown in Figure~\ref{fig:overview}, DischargeSim simulates multi-turn conversations between a DoctorAgent and a PatientAgent, structured around six clinically validated discharge topics: diagnosis, tests and treatments, return-to-hospital indicators, medication, post-discharge treatment, and follow-up. Each topic functions as a stage-specific education goal. The DoctorAgent must not only deliver accurate information but also apply appropriate strategies, such as simplification, emotional support, and shared decision-making, while adapting to the PatientAgent’s psychosocial profile (e.g., health literacy, education, emotional state).

Specifically, DischargeSim supports three interconnected evaluation layers:
\begin{enumerate}
    \item \textbf{Dialogue quality:} assessed using an LLM-as-Judge framework across dimensions of language and delivery, and human-centered communication. We further include evaluations of content completeness and strategy use, measuring whether the DoctorAgent delivers comprehensive information and applies appropriate communication strategies tailored to the patient profile.
    \item \textbf{Discharge summary generation:} evaluated using lexical metrics (ROUGE, BLEURT), medical metrics (UMLS-F1), and LLM-based criteria including fluency, personalization, and factuality.
    \item \textbf{Patient comprehension:} measured through expert-authored multiple-choice exams based on either the conversation or the generated summary.
\end{enumerate}

We evaluate 18 LLMs using DischargeSim and find that while larger models often produce more fluent responses, they do not consistently enhance patient comprehension, particularly when interacting with patients who have low health literacy or avoidant communication styles. Interestingly, some medium-sized models strike a better balance between strategy use and content completeness. Our key contributions include:
\begin{enumerate}
    \item introducing DischargeSim, the first benchmark for personalized discharge education through structured, multi-turn simulations;
    \item proposing a clinically grounded, three-stage evaluation framework covering dialogue quality, summary generation, and patient comprehension;
    \item benchmarking 18 LLMs and revealing unexpected behavioral trends such as strategy overuse and profile-sensitive performance gaps;
    \item presenting the first systematic analysis of how patient characteristics (e.g., health literacy, education, emotional state) affect model performance in discharge phase simulations, highlighting challenges in achieving equitable clinical AI communication.
\end{enumerate}

DischargeSim thus offers a novel testbed for advancing personalized, context-aware patient education beyond diagnostic reasoning.

\section{Methodology}

Our framework, DischargeSim, simulates the discharge process through a multi-agent interaction system that models post-visit patient--provider conversations. Unlike traditional diagnostic benchmarks which use static, single-turn QA formats, DischargeSim treats discharge communication as a sequential decision-making and educational task. The goal is not to infer a diagnosis, but to verify and enhance a patient's understanding of discharge instructions, including medications, follow-up care, activity restrictions, and warning signs.

\subsection{PatientAgent Profiles}

We configure PatientAgents with controlled variations in: (1) Health Literacy — low (limited health knowledge) vs. high (informed self-care); (2) Education Level — from no high school to bachelor’s degree; (3) Emotional Style — neutral, anxious, or deflective (avoidant) behavior.
These settings enable systematic evaluation of LLM adaptability, reasoning, and communication style across diverse patient profiles.
In addition, each PatientAgent is initialized from a Gold dataset that comprises 49 discharge notes sampled from the MIMIC-IV database~\cite{johnson_mimic-iv_2023}\footnote{Because this is MIMIC-IV data, only credentialed users can contact us to obtain the annotated Gold dataset. See \url{https://physionet.org/content/mimiciv/3.1/}.}.
For each note, domain experts manually created between 5 and 10 multiple-choice questions and answers.
The instructions and the detailed procedure for QA generation are included in the appendix~\ref{appendix:dataset_evaluation}.

\subsection{Data Preprocessing and Scenario Construction}

We utilize raw clinical data (free-text discharge notes and associated documentation) from the MIMIC-IV dataset. 
Each discharge note is parsed into a structured schema inspired by the OSCE framework~\cite{agentclinic2024} and AHRQ discharge guidelines~\cite{ahrq_going_home}, extracting: 1) Patient history and clinical course; 2) In-hospital test results and treatments; 3) Discharge diagnoses and medication plans; 4) Follow-up and return-to-hospital instructions.
To simulate patient memory constraints, we generate a parallel structured discharge representation that can be partially masked in PatientAgent profiles. 
The DoctorAgent has full access to the structured summary throughout the conversation.

\paragraph{Medical Content Stages}
\label{sec:Medical-Content}

DischargeSim dialogues are structured into six content stages~\cite{desai_empowering_2021}:
1) Return-to-Hospital Indicators: signs/symptoms that the patient should be aware of when they should contact or return to the hospital/Emergency Department.
2) Post-discharge Medications: the medications that the patient takes post-discharge.
3) Diagnosis Information: the chief complaint of the patient, the main and sub diagnoses of the patient. This should be in Unified Medical Language System (UMLS) vocabulary.
4) Activity Restrictions: what kind of actions or activities the patient should or should not be doing post-discharge.
5) In-hospital Treatments/Tests: what type of treatments/tests were done during their stay, and what the results were.
6) Follow-up Plans: when and where the patient should follow up the patient’s health issues post-discharge.
Each stage can span multiple turns and is initiated by the DoctorAgent.

\paragraph{Conversational Strategies}
\label{sec:strategy}

DoctorAgents employ six strategies~\cite{king_best_2013}:
1) Fostering relationship: build rapport and connection, respect patient statements, privacy, autonomy, engage in partnership building. Express caring and commitment. Use appropriate language. Encourage patient participation. Show interest in the patient as a person.  
2) Gathering information: attempt to understand the patient's needs for the encounter. Elicit full description of major reason for visit from biological and physiological perspectives. Ask open-ended questions. Allow patient to complete responses. Listen actively. Elicit patient's full set of concerns. Elicit patient's perspective on the problem/illness. Explore full effect of the illness. Clarify information. Inquire about additional concerns.  
3) Providing information: seek to understand patient's informational needs. Share information. Overcome barriers to patient understanding. Facilitate understanding. Explain nature of the problem and approach to diagnosis, treatment. Give uncomplicated explanations and instructions. Avoid jargon and complexity. Encourage questions and check understanding. Emphasize key messages.  
4) Decision making: outline collaborative action plan. Identify and enlist resources and support. Discuss follow-up and plan for unexpected outcomes.  
5) Enabling disease and treatment-related behavior: assess patient's interest in and capacity for self-management. Provide advice (information needs, coping skills, strategies for success). Agree on next steps. Assist patient to optimize autonomy and self-management. Arrange for needed support. Advocate for, and assist patient with, health system navigation. Assess patient's readiness to change health behaviors. Elicit goals, ideas, and decisions.  
6) Responding to emotions: facilitate patient expression of emotional consequences of illness. Acknowledge and explore emotions. Express empathy, sympathy, and reassurance. Provide help in dealing with emotions. Assess psychological distress.

\subsection{Simulation Workflow}

In DischargeSim, the discharge conversation is modeled as a multi-turn, multi-stage simulation between a DoctorAgent and a PatientAgent. Unlike traditional diagnosis-focused benchmarks that rely on single-turn QA formats, our framework reproduces post-visit clinical education in a structured and dynamic manner. This process comprises two core stages: interactive simulation and output generation.

\paragraph{Interactive Simulation Phase.} DoctorAgent leads a multi-turn conversation with the PatientAgent across six clinically critical content stages (as defined in \S~\ref{sec:Medical-Content}): return-to-hospital indicators, post-discharge medications, diagnosis information, activity restrictions, in-hospital treatments/tests, and follow-up plans. 
Each stage is initiated sequentially by the DoctorAgent and capped at 5 dialogue turns—except for the medications stage, which has no upper limit, to ensure complete coverage of all prescribed drugs. Responses by PatientAgent are influenced by its profile (education, health literacy, emotional state), and DoctorAgent is expected to adapt its language style and conversational strategy accordingly, referencing six strategy categories derived from AHRQ and medical communication literature (see \S~\ref{sec:strategy}).
We explicitly monitor whether DoctorAgent appropriately selects and applies these strategies—such as empathy, simplification, or clarification—to improve communication outcomes.

\paragraph{Post-conversation Output Generation.} After the dialogue, DoctorAgent is tasked with producing two personalized documents based on both the original discharge note and the full interaction history:
1) Personalized Discharge Summary: a free-text summary tailored not only to reflect accurate clinical information but also to emphasize content based on the PatientAgent’s cognitive and emotional profile. This extends traditional discharge summarization~\cite{cai2022generation} by incorporating patient-aware customization through conversational history.  
2) Structured AHRQ-style Checklist: a JSON-formatted structured discharge plan, modeled after the AHRQ guide~\cite{ahrq_going_home}. This checklist organizes the six discharge content areas into a machine-readable summary, allowing standardized comparisons with expert-written references.

\paragraph{Comprehension Exam for PatientAgent.} To assess patient education effectiveness, the PatientAgent takes a 10-question multiple-choice exam covering all six content categories. The questions are authored by clinical experts using the gold-standard discharge record and are designed to reflect the most essential information a patient should retain.
We evaluate comprehension under two input conditions to evaluate the support that patients might receive from Dialogue and generated Discharge Summary:
A) Dialogue Only: tests whether the conversation alone sufficiently conveys necessary information.  
B) Discharge Summary: mimics real-world scenarios where patients receive a written summary post-consultation.

\section{Experimental Design}

DischargeSim's evaluation framework focuses on three core components: (1) multi-turn dialogue quality, (2) personalized discharge summarization, and (3) patient comprehension outcomes.

\subsection{Dialogue Evaluation}

We adopt LLM-as-Judge, specifically using DeepSeek-V3~\cite{liu2024deepseek}, to evaluate each conversation across two main dimensions: Language and Delivery, and Human-Centered Communication. 
For Language and Delivery, we assess whether the language used is simple, fluent, and appropriate for the patient’s comprehension level (linguistic clarity), whether the response is logically structured and maintains a coherent, understandable flow (coherence), and whether it avoids unnecessary repetition that may hinder educational effectiveness (repetitiveness).
For Human-Centered Communication, we evaluate the extent to which the response demonstrates empathy and personalization by considering the patient's emotional state, health literacy, and context (personalization and empathy), as well as how appropriately the system responds to the patient's prior utterances, maintaining a natural and responsive conversational flow (interaction appropriateness). 
Check Table~\ref{fig:language-patient-eval} for more details.

\subsection{Discharge Summarization Evaluation}

After each interaction, the DoctorAgent generates a personalized discharge summary, which is evaluated using both automatic metrics and LLM-based judgment. 
We report ROUGE~\cite{lin_rouge_2004} to assess lexical overlap, BLEURT~\cite{sellam2020bleurt} to measure fluency, and UMLS-F1 to quantify clinical concept coverage.
In addition, we employ DeepSeek-V3-Judge to evaluate each summary across eight dimensions: 
semantic coverage (COV), which assesses how well the generated summary captures key semantic units from the reference;
factuality (FAC), which checks whether the summary maintains factual accuracy without hallucination; 
consistency (CON), which ensures internal coherence; 
informativeness (INF), which measures how concisely key ideas are conveyed;
coherence (COH), which evaluates logical flow and readability;
relevance (REL), which verifies that all content pertains to discharge instructions and follow-up care; 
fluency (FLU), which assesses grammatical and stylistic quality; 
and personalization (PER), which examines how well the summary is tailored to the patient’s literacy, emotional state, and specific clinical context. 
Check Table~\ref{fig:summ_eval_a} and \ref{fig:summ_eval_b} for more details.

\subsection{Validation of LLM-as-Judge Metrics}
\label{sec:llm-judge-validation}

To assess the reliability of our automatic evaluation framework, we conducted a human--LLM agreement study on 49 simulated discharge cases. Two licensed clinicians (one nurse and one physician) independently provided pairwise preferences between two nurse-agent outputs per case along five criteria: (i) Language Delivery, (ii) Human-Centered Communication, (iii) Content Coverage, (iv) Education Strategy, and (v) Overall Preference. Experts selected a preferred output and highlighted supporting evidence spans (green for positive, red for negative), which we used for qualitative error analysis.

Human Inter-rater Reliability.  
Agreement between the two experts was high: Cohen's $\kappa = 0.715$, percent agreement $= 85.7\%$, and Spearman's $\rho = 0.720$ ($p < 0.001$), indicating consistent expert judgments.

Alignment with LLM-as-Judge.  
We compared aggregated LLM-as-Judge ratings with expert preferences and observed substantial alignment: Cohen's $\kappa = 0.591$, percent agreement $= 79.6\%$, and Spearman's $\rho = 0.593$ ($p < 0.001$). These results support the use of LLM-as-Judge as a scalable proxy for expert assessment in discharge communication tasks.

\begin{table*}[!ht]
\resizebox{\textwidth}{!}{    
\begin{tabular}{l|c|c|ccccccc|ccccccc|cc}
\hline
& \textbf{Language} & \textbf{Human} & \multicolumn{7}{c|}{\textbf{Content Coverage}} & \multicolumn{7}{c|}{\textbf{Education Strategy}} & Exam & AHRQ \\

Model & \textbf{Delivery} & \textbf{Comm} & IRH & MED & DX & PDT & TDS & FU & avg. &
FR & GI & PI & DM & EBC & RE & avg. & ACC & ACC \\
\hline
qwen2.5-0.5b & \cellcolor{green!1}1.47 & \cellcolor{green!1}1.5 & \cellcolor{blue!1}2.79 & \cellcolor{blue!1}1.66 & \cellcolor{blue!1}2.44 & \cellcolor{blue!1}2.74 & \cellcolor{blue!1}1.4 & \cellcolor{blue!1}2.38 & \cellcolor{green!1}2.23 & \cellcolor{purple!1}2.24 & \cellcolor{purple!1}1.83 & \cellcolor{purple!1}2.53 & \cellcolor{purple!1}1.91 & \cellcolor{purple!1}2.31 & \cellcolor{purple!1}1.74 & \cellcolor{green!1}2.1 & \cellcolor{orange!1}52.87 & \cellcolor{orange!1}4.18 \\
qwen2.5-1.5b & \cellcolor{green!6}2.67 & \cellcolor{green!3}2.63 & \cellcolor{blue!6}3.3 & \cellcolor{blue!3}2.04 & \cellcolor{blue!11}4.24 & \cellcolor{blue!3}2.99 & \cellcolor{blue!3}2.36 & \cellcolor{blue!3}2.98 & \cellcolor{green!3}2.98 & \cellcolor{purple!3}3.11 & \cellcolor{purple!3}2.45 & \cellcolor{purple!3}3.58 & \cellcolor{purple!3}2.76 & \cellcolor{purple!3}3.14 & \cellcolor{purple!3}2.27 & \cellcolor{green!3}2.88 & \cellcolor{orange!8}62.54 & \cellcolor{orange!3}13.01 \\
qwen2.5-3b & \cellcolor{green!8}3.09 & \cellcolor{green!8}3.38 & \cellcolor{blue!8}4.05 & \cellcolor{blue!6}2.56 & \cellcolor{blue!16}4.47 & \cellcolor{blue!11}3.57 & \cellcolor{blue!13}3.12 & \cellcolor{blue!18}3.92 & \cellcolor{green!13}3.62 & \cellcolor{purple!8}3.55 & \cellcolor{purple!6}2.63 & \cellcolor{purple!8}3.9 & \cellcolor{purple!6}3.2 & \cellcolor{purple!8}3.64 & \cellcolor{purple!16}2.87 & \cellcolor{green!8}3.3 & \cellcolor{orange!11}64.95 & \cellcolor{orange!6}20.25 \\
qwen2.5-7b & \cellcolor{green!26}3.7 & \cellcolor{green!23}3.94 & \cellcolor{blue!13}4.09 & \cellcolor{blue!18}3.2 & \cellcolor{blue!13}4.35 & \cellcolor{blue!18}3.76 & \cellcolor{blue!16}3.43 & \cellcolor{blue!13}3.89 & \cellcolor{green!16}3.79 & \cellcolor{purple!18}3.76 & \cellcolor{purple!13}2.9 & \cellcolor{purple!13}4.07 & \cellcolor{purple!13}3.38 & \cellcolor{purple!13}3.71 & \cellcolor{purple!18}2.89 & \cellcolor{green!13}3.45 & \cellcolor{orange!16}66.77 & \cellcolor{orange!18}31.95 \\
qwen2.5-14b & \cellcolor{green!18}3.51 & \cellcolor{green!18}3.8 & \cellcolor{blue!26}4.31 & \cellcolor{blue!11}2.99 & \cellcolor{blue!18}4.49 & \cellcolor{blue!16}3.74 & \cellcolor{blue!21}3.57 & \cellcolor{blue!16}3.91 & \cellcolor{green!18}3.83 & \cellcolor{purple!31}4.0 & \cellcolor{purple!18}3.09 & \cellcolor{purple!21}4.21 & \cellcolor{purple!16}3.54 & \cellcolor{purple!18}3.91 & \cellcolor{purple!38}3.26 & \cellcolor{green!26}3.66 & \cellcolor{orange!13}66.47 & \cellcolor{orange!16}31.55 \\
qwen2.5-32b & \cellcolor{green!31}3.8 & \cellcolor{green!31}4.03 & \cellcolor{blue!21}4.23 & \cellcolor{blue!21}3.26 & \cellcolor{blue!21}4.59 & \cellcolor{blue!21}3.86 & \cellcolor{blue!23}3.75 & \cellcolor{blue!21}4.0 & \cellcolor{green!21}3.95 & \cellcolor{purple!33}4.01 & \cellcolor{purple!28}3.31 & \cellcolor{purple!28}4.33 & \cellcolor{purple!31}3.73 & \cellcolor{purple!23}3.99 & \cellcolor{purple!31}3.15 & \cellcolor{green!33}3.75 & \cellcolor{orange!21}69.79 & \cellcolor{orange!8}23.23 \\
qwen2.5-72b & \cellcolor{green!38}4.03 & \cellcolor{green!36}4.26 & \cellcolor{blue!23}4.27 & \cellcolor{blue!26}4.01 & \cellcolor{blue!28}4.77 & \cellcolor{blue!28}4.29 & \cellcolor{blue!33}4.35 & \cellcolor{blue!26}4.24 & \cellcolor{green!28}4.32 & \cellcolor{purple!41}4.16 & \cellcolor{purple!38}3.49 & \cellcolor{purple!41}4.43 & \cellcolor{purple!41}3.89 & \cellcolor{purple!41}4.16 & \cellcolor{purple!41}3.5 & \cellcolor{green!41}3.93 & \cellcolor{orange!28}74.02 & \cellcolor{orange!26}55.47 \\

\hline
llama3.2-1b & \cellcolor{green!3}2.64 & \cellcolor{green!6}2.78 & \cellcolor{blue!3}2.98 & \cellcolor{blue!13}3.11 & \cellcolor{blue!3}3.59 & \cellcolor{blue!6}3.06 & \cellcolor{blue!8}2.82 & \cellcolor{blue!6}3.51 & \cellcolor{green!6}3.18 & \cellcolor{purple!6}3.48 & \cellcolor{purple!8}2.88 & \cellcolor{purple!6}3.6 & \cellcolor{purple!8}3.24 & \cellcolor{purple!6}3.43 & \cellcolor{purple!21}3.03 & \cellcolor{green!6}3.28 & \cellcolor{orange!18}69.70 & \cellcolor{orange!11}25.13 \\
llama3.2-3b & \cellcolor{green!11}3.12 & \cellcolor{green!11}3.47 & \cellcolor{blue!11}4.08 & \cellcolor{blue!28}4.17 & \cellcolor{blue!23}4.60 & \cellcolor{blue!26}4.06 & \cellcolor{blue!26}3.93 & \cellcolor{blue!28}4.49 & \cellcolor{green!26}4.21 & \cellcolor{purple!16}3.63 & \cellcolor{purple!33}3.38 & \cellcolor{purple!16}4.08 & \cellcolor{purple!23}3.65 & \cellcolor{purple!21}3.93 & \cellcolor{purple!23}3.06 & \cellcolor{green!18}3.62 & \cellcolor{orange!23}71.00 & \cellcolor{orange!23}51.05 \\
llama3.1-8b & \cellcolor{green!13}3.32 & \cellcolor{green!13}3.66 & \cellcolor{blue!28}4.36 & \cellcolor{blue!33}4.83 & \cellcolor{blue!31}4.81 & \cellcolor{blue!36}4.51 & \cellcolor{blue!28}4.18 & \cellcolor{blue!33}4.67 & \cellcolor{green!31}4.56 & \cellcolor{purple!11}3.61 & \cellcolor{purple!21}3.17 & \cellcolor{purple!16}4.08 & \cellcolor{purple!18}3.58 & \cellcolor{purple!16}3.86 & \cellcolor{purple!13}2.86 & \cellcolor{green!16}3.52 & \cellcolor{orange!31}77.34 & \cellcolor{orange!33}77.50 \\
llama3.1-70b & \cellcolor{green!23}3.69 & \cellcolor{green!23}3.94 & \cellcolor{blue!36}4.48 & \cellcolor{blue!43}4.99 & \cellcolor{blue!36}4.84 & \cellcolor{blue!33}4.50 & \cellcolor{blue!31}4.24 & \cellcolor{blue!38}4.83 & \cellcolor{green!36}4.64 & \cellcolor{purple!26}3.92 & \cellcolor{purple!31}3.35 & \cellcolor{purple!26}4.30 & \cellcolor{purple!36}3.79 & \cellcolor{purple!28}4.02 & \cellcolor{purple!26}3.11 & \cellcolor{green!31}3.74 & \cellcolor{orange!41}84.29 & \cellcolor{orange!38}86.87 \\
llama3.3-70b & \cellcolor{green!43}4.13 & \cellcolor{green!36}4.26 & \cellcolor{blue!33}4.46 & \cellcolor{blue!36}4.94 & \cellcolor{blue!33}4.83 & \cellcolor{blue!31}4.42 & \cellcolor{blue!36}4.37 & \cellcolor{blue!36}4.69 & \cellcolor{green!33}4.62 & \cellcolor{purple!28}3.97 & \cellcolor{purple!43}3.54 & \cellcolor{purple!36}4.38 & \cellcolor{purple!38}3.88 & \cellcolor{purple!38}4.11 & \cellcolor{purple!33}3.17 & \cellcolor{green!38}3.83 & \cellcolor{orange!36}82.18 & \cellcolor{orange!36}86.51 \\

\hline
gpt3.5 & \cellcolor{green!21}3.59 & \cellcolor{green!16}3.72 & \cellcolor{blue!16}4.11 & \cellcolor{blue!16}3.18 & \cellcolor{blue!6}4.14 & \cellcolor{blue!8}3.39 & \cellcolor{blue!8}2.82 & \cellcolor{blue!11}3.69 & \cellcolor{green!11}3.56 & \cellcolor{purple!13}3.62 & \cellcolor{purple!8}2.88 & \cellcolor{purple!11}4.02 & \cellcolor{purple!11}3.30 & \cellcolor{purple!11}3.65 & \cellcolor{purple!6}2.74 & \cellcolor{green!11}3.37 & \cellcolor{orange!3}61.03 & \cellcolor{orange!21}32.30 \\
gpt4o & \cellcolor{green!36}3.98 & \cellcolor{green!33}4.16 & \cellcolor{blue!18}4.15 & \cellcolor{blue!23}3.91 & \cellcolor{blue!26}4.61 & \cellcolor{blue!23}3.93 & \cellcolor{blue!18}3.56 & \cellcolor{blue!23}4.06 & \cellcolor{green!23}4.03 & \cellcolor{purple!33}4.01 & \cellcolor{purple!36}3.43 & \cellcolor{purple!33}4.36 & \cellcolor{purple!33}3.76 & \cellcolor{purple!33}4.04 & \cellcolor{purple!28}3.13 & \cellcolor{green!36}3.78 & \cellcolor{orange!26}71.90 & \cellcolor{orange!28}58.22 \\
gpt4.1-nano & \cellcolor{green!16}3.48 & \cellcolor{green!18}3.80 & \cellcolor{blue!31}4.37 & \cellcolor{blue!8}2.58 & \cellcolor{blue!8}4.17 & \cellcolor{blue!13}3.62 & \cellcolor{blue!6}2.80 & \cellcolor{blue!8}3.63 & \cellcolor{green!8}3.53 & \cellcolor{purple!33}4.01 & \cellcolor{purple!26}3.27 & \cellcolor{purple!23}4.25 & \cellcolor{purple!26}3.67 & \cellcolor{purple!23}3.99 & \cellcolor{purple!36}3.19 & \cellcolor{green!28}3.72 & \cellcolor{orange!3}61.03 & \cellcolor{orange!13}26.90 \\
gpt4.1-mini & \cellcolor{green!33}3.92 & \cellcolor{green!41}4.27 & \cellcolor{blue!38}4.60 & \cellcolor{blue!38}4.96 & \cellcolor{blue!38}4.86 & \cellcolor{blue!38}4.62 & \cellcolor{blue!38}4.39 & \cellcolor{blue!31}4.66 & \cellcolor{green!38}4.68 & \cellcolor{purple!43}4.24 & \cellcolor{purple!41}3.50 & \cellcolor{purple!43}4.54 & \cellcolor{purple!43}3.94 & \cellcolor{purple!43}4.27 & \cellcolor{purple!43}3.62 & \cellcolor{green!43}4.01 & \cellcolor{orange!38}82.78 & \cellcolor{orange!41}87.43 \\
gpt5-nano & \cellcolor{green!28}3.73 & \cellcolor{green!28}4.01 & \cellcolor{blue!41}4.93 & \cellcolor{blue!31}4.75 & \cellcolor{blue!41}4.91 & \cellcolor{blue!41}4.90 & \cellcolor{blue!41}4.49 & \cellcolor{blue!41}4.95 & \cellcolor{green!41}4.82 & \cellcolor{purple!23}3.78 & \cellcolor{purple!23}3.18 & \cellcolor{purple!31}4.35 & \cellcolor{purple!28}3.70 & \cellcolor{purple!36}4.08 & \cellcolor{purple!8}2.82 & \cellcolor{green!23}3.65 & \cellcolor{orange!33}79.15 & \cellcolor{orange!31}73.80 \\
gpt5-mini & \cellcolor{green!41}4.07 & \cellcolor{green!43}4.28 & \cellcolor{blue!43}4.98 & \cellcolor{blue!41}4.97 & \cellcolor{blue!43}4.99 & \cellcolor{blue!43}4.99 & \cellcolor{blue!43}4.85 & \cellcolor{blue!43}4.99 & \cellcolor{green!43}4.96 & \cellcolor{purple!21}3.77 & \cellcolor{purple!16}3.06 & \cellcolor{purple!36}4.38 & \cellcolor{purple!21}3.64 & \cellcolor{purple!31}4.03 & \cellcolor{purple!8}2.82 & \cellcolor{green!18}3.62 & \cellcolor{orange!43}84.89 & \cellcolor{orange!43}93.80 \\

\hline
\end{tabular}
}
\caption{Evaluation of simulated discharge conversations across multiple dimensions:
\textbf{(1) Language \& Delivery} assesses linguistic clarity, coherence, and avoidance of repetitiveness.
\textbf{(2) Human-Centered Communication} evaluates personalization, empathy, and interaction appropriateness.
(Note: detailed scores for (1) and (2) are reported in the Appendix.)
\textbf{(3) Content Coverage} measures how well the conversation addresses key discharge information, including Indications to Return to Hospital/ED (IRH), Medication Information (MED), Diagnosis (DX), Post-Discharge Treatments (PDT), Treatments/Tests During Stay (TDS), and Follow-Up Instructions (FU). The average of these is reported as \textbf{Content Coverage avg.}.
\textbf{(4) Education Strategy} evaluates the use of effective communication strategies: Fostering Relationship (FR), Gathering Information (GI), Providing Information (PI), Decision Making (DM), Enabling Behavior Change (EBC), and Responding to Emotions (RE). The average is reported as \textbf{Strategy avg.}.
\textbf{(5) Exam ACC} measures the Patient Agent’s accuracy on comprehension questions when only the simulated conversation is used as input.
\textbf{(6) AHRQ ACC} compares the Doctor Agent-generated AHRQ checklist against the ground truth checklist based on the conversation.
Higher scores indicate better performance across all metrics.
}
\label{tab:conversation_results}
\end{table*}

% \paragraph{Human Evaluation}

% To complement automatic metrics and LLM-as-Judge assessments, we conducted detailed human evaluations on a subset of conversations and summaries.
% Each annotated case includes: (1) a gold-standard discharge summary, (2) a simulated doctor–patient conversation transcript, and (3) an LLM-generated discharge summary. To ensure unbiased judgment, model identities were anonymized and randomly shuffled across patient cases.
% Two domain-expert annotators were instructed to independently rate each case along multiple dimensions, including language quality, human-centered communication, discharge content accuracy, and education strategy.
% Summary quality was evaluated for fluency, factuality, relevance, coherence, and personalization.
% Annotators followed strict and conservative scoring guidelines to minimize inflation and promote reliable comparisons. 
% Full annotation guidelines are included in Table~\ref{tab:human-eval-annotation-guidelines}.

\subsection{Models}
We evaluate 18 LLMs, spanning commercial, open-source, and biomedical-specific variants: 
(1) OpenAI Series:  GPT-3.5-turbo, GPT-4o~\footnote{We don't include gpt-4o-mini-as-nurse because we use it as patient agents in our experiments.}, GPT-4.1 (incl. nano and mini), and GPT-5 (incl. nano and mini)
(2) Qwen 2.5 Series: 0.5B, 1.5B, 3B, 7B, 14B, 32B, 72B
(3) LLaMA3 Series: LLaMA3.1-8B/70B, LLaMA3.2-1B/3B, LLaMA3.3-70B
All models operate under unified prompting structures and profile settings aligned with DischargeSim's agent framework.

\section{Results}

We evaluate 18 large language models on both discharge conversation (Table~\ref{tab:conversation_results}) and discharge summary generation tasks (Table~\ref{tab:summary_results}), spanning language fluency, human-centeredness, medical coverage, strategy use, and downstream patient comprehension.

\subsection{Language \& Communication.}
We observe clear scaling trends. The highest Language \& Delivery / Human-Centered Communication scores appear in LLaMA3.3-70B (4.13 / 4.26) and GPT-5-mini (4.07 / 4.28), with Qwen2.5-72B (4.03 / 4.26) and GPT-4.1-mini (3.92 / 4.27) close behind. GPT-5-nano is also strong (3.73 / 4.01), though slightly below the top tier. In contrast, Qwen2.5-0.5B is markedly lower (1.47 / 1.50), reflecting limited fluency and engagement.

\subsection{Content Coverage.}
Top models deliver broad and consistent coverage across discharge topics. The best average coverage scores are achieved by GPT-5-mini (4.96), GPT-5-nano (4.82), GPT-4.1-mini (4.68), LLaMA3.1-70B (4.64), and LLaMA3.3-70B (4.62). On key categories, several models reach high marks on MED and DX—for example, GPT-4.1-mini MED 4.96 / DX 4.86 and LLaMA3.3-70B MED 4.94 / DX 4.83—while the GPT-5 series approaches ceiling across most items (e.g., GPT-5-mini IRH 4.98 / MED 4.97 / DX 4.99 / FU 4.99).

\subsection{Education Strategy Use.}
Aggregate strategy use peaks with GPT-4.1-mini (strategy avg. 4.01), followed by Qwen2.5-72B (3.93), LLaMA3.3-70B (3.83), and LLaMA3.1-70B (3.74). The GPT-5 models are slightly lower on aggregate (mini 3.62; nano 3.65), with relatively weaker scores in Responding to Emotions and Gathering Information (e.g., mini RE 2.82, GI 3.06). Lower-tier models underuse Responding to Emotions and Enabling Behavior Change (e.g., Qwen2.5-0.5B RE 2.10, EBC 2.31), indicating challenges in nuanced patient communication.

\subsection{Patient Comprehension Outcomes (Conversations).}
Exam accuracy (PatientAgent using only the conversation) rises with model quality: GPT-5-mini 84.89\%, LLaMA3.1-70B 84.29\%, GPT-4.1-mini 82.78\%, LLaMA3.3-70B 82.18\%, with GPT-5-nano also high at 79.15\%. In the Qwen series, accuracy generally increases with size (52.87\% → 74.02\%), with a slight dip at 14B (66.47\%). AHRQ checklist accuracy shows a similar trend, topping out at GPT-5-mini 93.80\% and strong open-source results for LLaMA3.1-70B (86.87\%) and LLaMA3.3-70B (86.51\%).

\begin{table*}[!ht]
\resizebox{\textwidth}{!}{ 
\begin{tabular}{l|ccc|ccccc|ccccc|c}
\hline
& \multicolumn{3}{c|}{\textbf{Generation Metrics}} & 
 \multicolumn{5}{c|}{\textbf{Language \& Appropriateness}} & \multicolumn{5}{c|}{\textbf{Factuality \& Completeness}} & Exam  \\
Model & RougeL & BLEURT & UMLS-F & FL & CO & INF & PER & avg. & SC & FAC & REL & CON & avg. & ACC \\
\hline
qwen2.5-0.5b & \cellcolor{blue!8}0.1530 & \cellcolor{blue!1}0.3163 & \cellcolor{blue!1}0.1524 & \cellcolor{DoctorColor!1}3.09 & \cellcolor{DoctorColor!1}2.74 & \cellcolor{DoctorColor!1}2.79 & \cellcolor{DoctorColor!1}2.65 & \cellcolor{DoctorColor!1}2.82 & \cellcolor{PatientColor!1}1.96 & \cellcolor{PatientColor!1}1.63 & \cellcolor{PatientColor!1}2.93 & \cellcolor{PatientColor!1}1.59 & \cellcolor{PatientColor!1}2.02 & \cellcolor{cyan!1}52.87\% \\
qwen2.5-1.5b & \cellcolor{blue!1}0.1152 & \cellcolor{blue!3}0.3188 & \cellcolor{blue!3}0.1764 & \cellcolor{DoctorColor!3}3.28 & \cellcolor{DoctorColor!4}2.97 & \cellcolor{DoctorColor!3}2.86 & \cellcolor{DoctorColor!3}2.87 & \cellcolor{DoctorColor!3}3.00 & \cellcolor{PatientColor!6}2.48 & \cellcolor{PatientColor!6}2.54 & \cellcolor{PatientColor!6}3.56 & \cellcolor{PatientColor!6}2.60 & \cellcolor{PatientColor!6}2.79 & \cellcolor{cyan!6}57.10\% \\
qwen2.5-3b & \cellcolor{blue!16}0.1897 & \cellcolor{blue!8}0.3608 & \cellcolor{blue!11}0.2409 & \cellcolor{DoctorColor!8}3.73 & \cellcolor{DoctorColor!12}3.73 & \cellcolor{DoctorColor!23}3.58 & \cellcolor{DoctorColor!11}3.50 & \cellcolor{DoctorColor!13}3.64 & \cellcolor{PatientColor!13}3.11 & \cellcolor{PatientColor!11}2.94 & \cellcolor{PatientColor!8}4.27 & \cellcolor{PatientColor!11}3.26 & \cellcolor{PatientColor!8}3.39 & \cellcolor{cyan!16}63.44\% \\
qwen2.5-7b & \cellcolor{blue!18}0.2018 & \cellcolor{blue!11}0.3718 & \cellcolor{blue!21}0.2834 & \cellcolor{DoctorColor!16}3.94 & \cellcolor{DoctorColor!19}3.87 & \cellcolor{DoctorColor!13}3.50 & \cellcolor{DoctorColor!11}3.50 & \cellcolor{DoctorColor!16}3.70 & \cellcolor{PatientColor!16}3.16 & \cellcolor{PatientColor!8}2.93 & \cellcolor{PatientColor!13}4.43 & \cellcolor{PatientColor!8}3.18 & \cellcolor{PatientColor!11}3.41 & \cellcolor{cyan!13}63.14\% \\
qwen2.5-14b & \cellcolor{blue!13}0.1868 & \cellcolor{blue!18}0.3880 & \cellcolor{blue!16}0.2767 & \cellcolor{DoctorColor!31}4.02 & \cellcolor{DoctorColor!25}3.90 & \cellcolor{DoctorColor!16}3.51 & \cellcolor{DoctorColor!21}3.54 & \cellcolor{DoctorColor!21}3.74 & \cellcolor{PatientColor!18}3.24 & \cellcolor{PatientColor!18}3.50 & \cellcolor{PatientColor!21}4.73 & \cellcolor{PatientColor!21}3.76 & \cellcolor{PatientColor!18}3.80 & \cellcolor{cyan!11}62.84\% \\
qwen2.5-32b & \cellcolor{blue!11}0.1830 & \cellcolor{blue!23}0.3980 & \cellcolor{blue!13}0.2674 & \cellcolor{DoctorColor!18}3.98 & \cellcolor{DoctorColor!22}3.89 & \cellcolor{DoctorColor!11}3.48 & \cellcolor{DoctorColor!16}3.51 & \cellcolor{DoctorColor!18}3.72 & \cellcolor{PatientColor!21}3.28 & \cellcolor{PatientColor!21}3.62 & \cellcolor{PatientColor!18}4.71 & \cellcolor{PatientColor!18}3.75 & \cellcolor{PatientColor!21}3.83 & \cellcolor{cyan!21}64.35\% \\
qwen2.5-72b & \cellcolor{blue!33}0.2545 & \cellcolor{blue!31}0.4089 & \cellcolor{blue!33}0.3709 & \cellcolor{DoctorColor!33}4.04 & \cellcolor{DoctorColor!38}3.97 & \cellcolor{DoctorColor!26}3.62 & \cellcolor{DoctorColor!26}3.57 & \cellcolor{DoctorColor!28}3.80 & \cellcolor{PatientColor!28}3.77 & \cellcolor{PatientColor!26}4.06 & \cellcolor{PatientColor!28}4.86 & \cellcolor{PatientColor!28}4.22 & \cellcolor{PatientColor!28}4.22 & \cellcolor{cyan!28}73.41\% \\
\hline
llama3.2-1b & \cellcolor{blue!21}0.2059 & \cellcolor{blue!6}0.3325 & \cellcolor{blue!6}0.1909 & \cellcolor{DoctorColor!6}3.51 & \cellcolor{DoctorColor!6}3.01 & \cellcolor{DoctorColor!6}3.28 & \cellcolor{DoctorColor!6}3.19 & \cellcolor{DoctorColor!6}3.25 & \cellcolor{PatientColor!3}2.24 & \cellcolor{PatientColor!3}1.88 & \cellcolor{PatientColor!3}3.19 & \cellcolor{PatientColor!3}1.79 & \cellcolor{PatientColor!3}2.27 & \cellcolor{cyan!18}63.64\% \\
llama3.2-3b & \cellcolor{blue!26}0.2225 & \cellcolor{blue!26}0.4025 & \cellcolor{blue!23}0.3166 & \cellcolor{DoctorColor!11}3.76 & \cellcolor{DoctorColor!9}3.53 & \cellcolor{DoctorColor!28}3.63 & \cellcolor{DoctorColor!28}3.60 & \cellcolor{DoctorColor!11}3.63 & \cellcolor{PatientColor!23}3.49 & \cellcolor{PatientColor!23}3.68 & \cellcolor{PatientColor!23}4.80 & \cellcolor{PatientColor!23}3.93 & \cellcolor{PatientColor!23}3.96 & \cellcolor{cyan!23}65.26\% \\
llama3.1-8b & \cellcolor{blue!38}0.2741 & \cellcolor{blue!36}0.4285 & \cellcolor{blue!36}0.3818 & \cellcolor{DoctorColor!18}3.98 & \cellcolor{DoctorColor!17}3.80 & \cellcolor{DoctorColor!33}3.80 & \cellcolor{DoctorColor!33}3.67 & \cellcolor{DoctorColor!33}3.81 & \cellcolor{PatientColor!31}4.02 & \cellcolor{PatientColor!31}4.30 & \cellcolor{PatientColor!31}4.98 & \cellcolor{PatientColor!31}4.36 & \cellcolor{PatientColor!31}4.41 & \cellcolor{cyan!31}76.44\% \\
llama3.1-70b & \cellcolor{blue!43}0.3419 & \cellcolor{blue!43}0.4535 & \cellcolor{blue!43}0.4251 & \cellcolor{DoctorColor!36}4.09 & \cellcolor{DoctorColor!30}3.92 & \cellcolor{DoctorColor!36}3.86 & \cellcolor{DoctorColor!36}3.69 & \cellcolor{DoctorColor!36}3.89 & \cellcolor{PatientColor!41}4.33 & \cellcolor{PatientColor!38}4.62 & \cellcolor{PatientColor!43}5.00 & \cellcolor{PatientColor!36}4.66 & \cellcolor{PatientColor!38}4.65 & \cellcolor{cyan!41}82.48\% \\
llama3.3-70b & \cellcolor{blue!41}0.2865 & \cellcolor{blue!41}0.4463 & \cellcolor{blue!41}0.4198 & \cellcolor{DoctorColor!23}4.01 & \cellcolor{DoctorColor!27}3.91 & \cellcolor{DoctorColor!31}3.68 & \cellcolor{DoctorColor!31}3.61 & \cellcolor{DoctorColor!28}3.80 & \cellcolor{PatientColor!33}4.05 & \cellcolor{PatientColor!33}4.46 & \cellcolor{PatientColor!38}4.99 & \cellcolor{PatientColor!33}4.53 & \cellcolor{PatientColor!33}4.50 & \cellcolor{cyan!38}80.36\% \\
\hline
gpt3.5 & \cellcolor{blue!6}0.1504 & \cellcolor{blue!16}0.3874 & \cellcolor{blue!18}0.2790 & \cellcolor{DoctorColor!13}3.92 & \cellcolor{DoctorColor!14}3.74 & \cellcolor{DoctorColor!8}3.29 & \cellcolor{DoctorColor!8}3.36 & \cellcolor{DoctorColor!8}3.58 & \cellcolor{PatientColor!8}2.99 & \cellcolor{PatientColor!13}3.33 & \cellcolor{PatientColor!11}4.40 & \cellcolor{PatientColor!13}3.40 & \cellcolor{PatientColor!13}3.52 & \cellcolor{cyan!8}59.52\% \\
gpt4o & \cellcolor{blue!23}0.2076 & \cellcolor{blue!33}0.4149 & \cellcolor{blue!26}0.3598 & \cellcolor{DoctorColor!23}4.01 & \cellcolor{DoctorColor!35}3.93 & \cellcolor{DoctorColor!21}3.56 & \cellcolor{DoctorColor!21}3.54 & \cellcolor{DoctorColor!26}3.76 & \cellcolor{PatientColor!26}3.55 & \cellcolor{PatientColor!28}4.12 & \cellcolor{PatientColor!26}4.83 & \cellcolor{PatientColor!26}4.18 & \cellcolor{PatientColor!26}4.16 & \cellcolor{cyan!26}67.98\% \\
gpt4.1-nano & \cellcolor{blue!3}0.1299 & \cellcolor{blue!13}0.3738 & \cellcolor{blue!8}0.2322 & \cellcolor{DoctorColor!23}4.01 & \cellcolor{DoctorColor!30}3.92 & \cellcolor{DoctorColor!18}3.52 & \cellcolor{DoctorColor!18}3.52 & \cellcolor{DoctorColor!21}3.74 & \cellcolor{PatientColor!11}3.09 & \cellcolor{PatientColor!13}3.33 & \cellcolor{PatientColor!16}4.51 & \cellcolor{PatientColor!16}3.55 & \cellcolor{PatientColor!16}3.61 & \cellcolor{cyan!3}56.19\% \\
gpt4.1-mini & \cellcolor{blue!36}0.2653 & \cellcolor{blue!38}0.4445 & \cellcolor{blue!38}0.3954 & \cellcolor{DoctorColor!41}4.18 & \cellcolor{DoctorColor!40}4.07 & \cellcolor{DoctorColor!41}4.02 & \cellcolor{DoctorColor!38}3.88 & \cellcolor{DoctorColor!41}4.04 & \cellcolor{PatientColor!38}4.27 & \cellcolor{PatientColor!41}4.82 & \cellcolor{PatientColor!31}4.98 & \cellcolor{PatientColor!41}4.81 & \cellcolor{PatientColor!41}4.72 & \cellcolor{cyan!36}79.76\% \\
gpt5-nano & \cellcolor{blue!28}0.2341 & \cellcolor{blue!28}0.4076 & \cellcolor{blue!28}0.3638 & \cellcolor{DoctorColor!38}4.12 & \cellcolor{DoctorColor!43}4.08 & \cellcolor{DoctorColor!38}4.00 & \cellcolor{DoctorColor!41}3.91 & \cellcolor{DoctorColor!38}4.03 & \cellcolor{PatientColor!36}4.24 & \cellcolor{PatientColor!36}4.56 & \cellcolor{PatientColor!31}4.98 & \cellcolor{PatientColor!36}4.66 & \cellcolor{PatientColor!36}4.61 & \cellcolor{cyan!33}77.04\% \\
gpt5-mini & \cellcolor{blue!31}0.2451 & \cellcolor{blue!21}0.3919 & \cellcolor{blue!31}0.3663 & \cellcolor{DoctorColor!43}4.24 & \cellcolor{DoctorColor!43}4.08 & \cellcolor{DoctorColor!43}4.31 & \cellcolor{DoctorColor!43}4.13 & \cellcolor{DoctorColor!43}4.19 & \cellcolor{PatientColor!43}4.48 & \cellcolor{PatientColor!43}4.97 & \cellcolor{PatientColor!38}4.99 & \cellcolor{PatientColor!43}4.93 & \cellcolor{PatientColor!43}4.84 & \cellcolor{cyan!43}83.38\% \\

\hline
\end{tabular}
}
\caption{Evaluation of LLM-generated discharge summaries. The table reports results across three metric categories: (1) \textbf{Traditional Generation Evaluation Metrics} including ROUGE-L, BLEURT, UMLS-F, and Exam Accuracy; (2) LLM-Judge Evaluation Criteria for \textbf{Language Quality \& Appropriateness} covering Fluency (FL), Coherence (CO), Informativeness (INF), and Personalization (PER); and (3) LLM-Judge Evaluation Criteria for \textbf{Evidence-Based Factuality \& Completeness} covering Semantic Coverage (SC), Factuality (FAC), Relevance (REL), and Consistency (CON). We also report Exam ACC, which measures the accuracy of the Patient Agent on comprehension questions when only the discharge summary is provided as input. Higher scores indicate better performance.
}
\label{tab:summary_results}
\end{table*}

\subsection{Discharge Summary Generation.}
For summaries, LLaMA3.1-70B leads the lexical/clinical overlap metrics (ROUGE-L 0.3419, UMLS-F 0.4251). GPT-5-mini attains the highest LLM-judged quality and factuality (L\&A avg. 4.19; F\&C avg. 4.84) and the best downstream comprehension (83.38\%), with GPT-5-nano also strong (L\&A avg. 4.03; F\&C avg. 4.61; 77.04\%). LLaMA3.3-70B is close on judged factuality/completeness (avg. 4.50), and GPT-4.1-mini shows excellent judged quality (L\&A avg. 4.04; F\&C avg. 4.72) despite a lower ROUGE-L (0.2653). Lower-capacity models, such as Qwen2.5-3B and LLaMA3.2-1B, yield lower summary-only comprehension (63–64\%).

\subsection{Overall Ranking and Trends.}
Among open-source models, LLaMA3.1-70B and LLaMA3.3-70B provide the strongest overall balance across conversation and summary tasks. Qwen2.5-72B is competitive on language/human scores (4.03 / 4.26) but trails on coverage (avg. 4.32), strategy (3.93), and conversation exam accuracy (74.02\%) relative to GPT-4.1-mini (4.68 / 4.01 / 82.78\%). GPT-5-mini generally sets the upper bound on coverage, exam, and AHRQ in conversations (4.96 / 84.89\% / 93.80\%) and leads judged quality/factuality in summaries (4.19 / 4.84), while GPT-5-nano offers a competitive but slightly lower profile.

\section{Discussion}

\subsection{Impact of Patient Health Literacy}

To investigate how patient health literacy affects dialogue agent behavior, we compare model responses in two scenarios: HL1 (interacting with patients of low health literacy) and HL2 (interacting with patients of high health literacy). Figure~\ref{fig:HL_diff} visualizes the relative difference across six evaluation dimensions, computed as $(\text{HL2} - \text{HL1}) / \text{HL1}$.

Overall, high-literacy patients (HL2) consistently elicit better responses across most models and dimensions. Improvements are especially pronounced in Language and Delivery, Human-Centered Communication, and Strategy, indicating that models are more capable of producing fluent, coherent, and empathetic responses when provided with clearer, more structured patient input. In contrast, factuality, completeness, language, and Appropriateness remain stable, indicating robustness to patient comprehension levels.

As shown in Figure~\ref{fig:HL_diff_criteria}, these expressive dimensions also exhibit the highest variability across models, indicating that they are more sensitive to the user's literacy level.
We further investigate this effect within the Qwen model family (excluding the outlier Qwen-0.5B). Figure~\ref{fig:HL_diff_modelsize} shows a strong positive correlation between model size and the average performance gap between HL2 and HL1 ($r=0.80$, $p=0.053$), suggesting that larger models are better at tailoring their responses to high-literacy patients. This effect is most significant in Language and Delivery ($r=0.97$, $p<0.01$) and Human-Centered Communication ($r=0.86$, $p<0.05$), as shown in Figure~\ref{fig:HL_diff_modelsize_criteria}.
These findings underscore the need for future research to enhance model robustness in low-literacy settings. In particular, methods to enhance strategy generation and communication clarity should be prioritized to ensure equitable support for diverse patient populations.

\subsection{Impact of Patient Education Level}

To investigate how patient education level affects dialogue agent performance, we compare responses under three conditions: Education 1 (no high school education), Education 2 (high school/GED), and Education 3 (college graduate). For each model and evaluation dimension, we compute the relative score difference (e.g., $(\text{Edu2} - \text{Edu1}) / \text{Edu1}$) to quantify sensitivity to patient educational background.

We observe a consistent improvement in agent performance as patient education level increases. Compared to Education 1, both Education 2 and Education 3 users elicit notably better responses across expressive dimensions, such as Strategy, Human-Centered Communication, and Language and Delivery. These improvements suggest that higher education levels, which likely correlate with more precise phrasing and more structured conversational behavior, enable agents to deliver more coherent and empathetic responses. In contrast, dimensions such as Factuality, Completeness, Language, and Appropriateness exhibit minimal variation, indicating that these aspects are more dependent on the model’s internal knowledge and less influenced by the quality of user input.
Figure~\ref{fig:edu_diff_3in1} shows the relative differences across all models for three education level comparisons.

We also measure the variability of these differences across models for each dimension. As shown in Figure~\ref{fig:edu_diff_criteria}, expressive and interaction-heavy aspects such as Strategy and Content demonstrate the highest variation across models, while factual and linguistic dimensions remain more stable.
Focusing on the Qwen model family (excluding the 0.5B outlier), we find significant correlations between model size and education sensitivity across dimensions. Larger models, such as Qwen-14B and Qwen-72B, are more responsive to users with higher education, especially in expressive dimensions like Language and Delivery, and Human-Centered Communication (Figure~\ref{fig:edu_diff_modelsize}).

These findings emphasize the importance of education-aware agent design. While large models are generally more capable of tailoring responses to educated users, performance equity must be ensured for users with limited formal education. Future work should explore controllable generation techniques and robust fine-tuning strategies to enable high-quality communication regardless of educational background.

\subsection{Impact of Patient Emotional Style}
\label{sec:emotion-style}

To assess how patient emotional style influences dialogue agent performance, we analyze responses under three simulated conditions: deflective, neutral, and anxious. For each model and evaluation dimension, we compute the relative score difference (e.g., (anxious - deflective)/deflective) to quantify the variation in agent responses across emotional inputs.

Overall, we observe that anxious patients consistently elicit more supportive and strategic responses, particularly in dimensions such as Human-Centered Communication, Language and Delivery, and Strategy. This suggests that LLM-based agents often adapt their tone and conversational strategies when confronted with emotionally charged user behavior. By contrast, dimensions such as Language, Appropriateness, Factuality, and Completeness remain relatively stable across emotional styles, indicating that factual correctness and linguistic formality are governed more by the model’s internal representations than by emotional cues from the patient. The trend across models is visualized in Figure~\ref{fig:emo_diff_trio}.

We further compute the standard deviation of these score differences across models to assess robustness. As shown in Figure~\ref{fig:emo_diff_criteria}, Language and Delivery and Human-Centered Communication exhibit the highest variability, suggesting they are most sensitive to patient emotional style. In contrast, Factuality and Appropriateness remain stable, reinforcing their input-agnostic nature.
Zooming in on the Qwen model family (excluding the 0.5B outlier), we find that larger models are more robust and less reactive to variations in emotional style. In the Anxious vs. Neutral comparison, model size is strongly negatively correlated with performance sensitivity in expressive dimensions: Strategy ($r = -0.91$), Language and Appropriateness ($r = -0.73$), and Factuality and Completeness ($r = -0.64$), as shown in Figure~\ref{fig:emo_diff_modelsize}. 

This suggests that small models tend to overreact to emotional input, while larger models exhibit more calibrated responses. In the Neutral vs. Deflective contrast, correlations turn moderately positive in Human-Centered Communication ($r = 0.45$) and Content ($r = 0.59$), indicating that larger models are more effective at recognizing and adapting to avoidant conversational behavior.

\section{Related Work}

Effective discharge communication is a well-established determinant of patient adherence and recovery, yet it remains underexplored in NLP benchmarks~\cite{yao2025survey}. Prior studies have highlighted that patients, particularly those with low health literacy or emotional distress, often struggle to retain critical discharge information~\cite{richard2017communication, weerahandi2018predictors}. Although clinical best practices such as teach-back, simplified language, and checklist-based summaries (e.g., AHRQ’s “Going Home” guide) are known to enhance patient understanding, these strategies are seldom modeled or evaluated in current LLM-based systems. Studies consistently indicate that between 40\% and 80\% of patients forget or misunderstand critical discharge instructions shortly after hospital visits~\cite{kessels2003patients, jack2009reengineered}, underscoring the need for communication-centered evaluation.

Existing clinical dialogue benchmarks, such as AgentClinic~\cite{agentclinic2024}, AMIE~\cite{tu2025towards}, and HealthBench~\cite{arora2025healthbench}, primarily focus on in-visit diagnostic conversations, emphasizing uncertainty reasoning, evidence integration, and stepwise clinical inference. Benchmarks like MedAgents~\cite{tang2023medagents} further simulate multimodal physician–patient interactions for sequential decision-making. However, these works leave the crucial post-visit discharge phase unaddressed, where the primary goal shifts from diagnostic accuracy to communication clarity and patient-centered education. Similarly, domain-specific medical LLM systems such as ChatDoctor~\cite{li2023chatdoctor}, NoteChat~\cite{wang2023notechat}, PMC-LLaMA~\cite{wu2024pmc}, MedAlpaca~\cite{han2023medalpaca}, BioInstruct~\cite{tran2024bioinstruct}, and Huatuogpt~\cite{zhang2023huatuogpt} have primarily focused on diagnostic QA, knowledge recall, and retrieval-augmented dialogue, typically evaluating factual correctness or reasoning accuracy in single-turn settings. In contrast, DischargeSim targets multi-turn, post-visit educational conversations grounded in patient-specific profiles (literacy, emotion, comprehension level), using downstream comprehension testing as a core evaluation signal, dimensions underrepresented in prior medical AI evaluation frameworks.

While PaniniQA~\cite{paniniqa2023} marks initial progress by evaluating patient comprehension through QA on discharge summaries, it remains limited to static, single-turn question-answer formats and overlooks crucial discharge tasks such as personalized summary generation and structured AHRQ checklist completion. Dialogue-based educational interventions have long shown advantages in enhancing comprehension and learning outcomes, particularly among populations with lower health literacy~\cite{whitehurst2002dialogic, golinkoff2019conversational}. However, NLP approaches to dialogue modeling and educational QA~\cite{du2017learning, shwartz2020unsupervised} have yet to be systematically applied to discharge education in a way that accounts for patient diversity and dynamic adaptation. DischargeSim systematically integrates multi-turn, personalized educational dialogues into clinical discharge settings, extending beyond QA to encompass personalized summary generation and checklist-based instruction, while explicitly modeling psychosocial and cognitive patient factors.

In addition, inspired by advances in LLM-as-Judge research, DischargeSim incorporates tailored evaluation rubrics for distinct patient-centered communication tasks. In the broader NLP community, LLMs such as GPT-4~\cite{achiam2023gpt, liu2023gpteval, fu2023gptscore} and critique-tuned variants~\cite{ke2023critiquellm} have been applied as automated evaluators for summarization~\cite{chen2023storyer}, dialogue~\cite{zheng2024judging, zhang2024comprehensive}, and translation~\cite{kocmi2023large}. This paradigm has recently been extended to the medical domain for evaluating clinical conversations~\cite{tu2025towards, arora2025healthbench}, medical documentation~\cite{croxford2025automating, chung2025verifact, brake2024comparing}, exam QA~\cite{yao2024mcqg, yao2024medqa}, and reasoning~\cite{jeong2024improving, tran2024rare}. DischargeSim builds upon this body of work by offering a structured, task-specific judging framework tailored to discharge communication, enabling robust, scalable, and context-aware evaluation of LLMs in patient education.

\section{Conclusion}
DischargeSim presents a simulation framework for evaluating LLMs in personalized, post-visit patient education. Our results show significant variation in performance across communication strategies, content coverage, and comprehension support, reflecting sensitivity to patient literacy, education, and emotional state. This work lays the foundation for more equitable, reliable, and patient-centered AI systems in clinical communication.

\section{Limitations}
While DischargeSim introduces a novel framework for evaluating LLMs in post-visit patient education, it has several limitations.

First, our PatientAgent simulation relies on scripted profiles and LLM-generated responses, which, despite being carefully constructed, may not fully capture the variability and unpredictability of real-world patient behavior. Incorporating human-in-the-loop evaluations or standardized patient actors could yield more realistic interactions.

Second, although we evaluate over 18 LLMs, our benchmark focuses primarily on English-language discharge scenarios derived from the MIMIC-IV dataset. The generalizability of our findings to other languages, health systems, and discharge formats (e.g., pediatrics, oncology) remains an open question.

Third, our evaluation pipeline, while comprehensive, still relies on LLM-as-Judge assessments and multiple-choice comprehension tests. While these offer scalable and interpretable insights, they may miss nuanced qualitative aspects of patient understanding and trust. Future work should integrate real patient feedback and long-term behavioral outcomes.

Finally, our current framework emphasizes structured discharge topics and may not fully account for free-form or emotion-driven interactions that commonly occur during discharge discussions. Our benchmark currently structures conversations around six predefined discharge topics derived from widely recognized clinical guidelines (diagnosis, medications, post-discharge activity, return indicators, in-hospital tests, and follow-up). While this ensures consistent coverage of essential safety information, it may constrain the natural, patient-driven flow of real-world discharge dialogues. Actual interactions often evolve dynamically, with patients steering the conversation toward personally salient concerns. In future iterations, we plan to incorporate flexible topic prioritization and adaptive dialogue management strategies that preserve critical safety coverage while enabling models to respond more fluidly to individual patient cues and evolving needs.

\section{Ethics Statement}
This study presents a simulation framework designed to evaluate and improve the patient education capabilities of large language models in discharge communication scenarios. All data used in this study are derived from the publicly available MIMIC-IV dataset, which is fully de-identified and widely adopted in clinical NLP research. No personally identifiable information is included.

The comprehension exam dataset was created through a two-step process. First, multiple-choice questions were drafted by PhD students in Computer Science based in the United States, with each questionnaire containing 5--10 questions comprising an answer, a distractor, and an irrelevant option. These questions were then reviewed and revised by three medical experts, including two nursing professors from the United States and one physician from South Korea, who validated the quality of the questions and provided comments or edits to ensure medical accuracy and clarity.

All medical experts were compensated for their time at a rate of \$40 per hour, in line with academic ethics standards. The annotation and review process was conducted under an approved academic protocol to ensure transparency and quality control.

DischargeSim is intended solely for research and evaluation purposes and has not been clinically validated for real-world deployment. Given the sensitivity of discharge communication, we caution against using LLM-based systems in clinical settings without expert oversight. We strongly advocate for continued research into the safety, fairness, and interpretability of AI systems in healthcare to support equitable and responsible integration into clinical workflows.

% maybe appenxi this table, how the conent and strategy work

\section*{Acknowledgments}

This material is the result of work supported with resources and the use of facilities at the Center for Healthcare Organization and Implementation Research, VA Bedford Health Care.

\bibliography{custom}

\appendix
\label{sec:appendix}

\section{Data Creation}
\label{appendix:dataset_evaluation}

We took two steps to generate and evaluate the quality of Comprehension Exam questions.
First, we asked students to annotate 5-10 questionnaires.
Then, 3 experts went through the generated datasets and evaluated, commented on, or modified the questionnaires.
The students were all PhD students majoring in Computer Science in the United States. 
The expert annotators were 2 nurse professors in the United States and 1 doctor from South Korea. 
For each question, we instructed the annotators to generate 5 to 10 multiple-choice questions with three choices: answer, distractor, and irrelevant, as shown in Figure~\ref{fig:guideline-generation-for-Q}.
We asked three medical experts to go through the questions and questionnaires to validate the quality of the annotations. 
We asked them to modify or leave comments and made the changes according to their comments (Figure~\ref{fig:silver-expert-annotation-guideline}).

\begin{figure*}
    \centering
\begin{tcolorbox}[width=\linewidth,
                  % enhanced,
                  % sharp corners,
                  %%frame hidden,
                  % interior hidden,
                  % halign=flush left,
                  boxsep=0pt,
                  left=5pt,
                  right=5pt,
                  top=5pt,
                  title=Guidelines for Annotation,
                  ]%%
% \centering

1. You are going to create 5-10 questions for each discharge note. \\
2. These questions are going to be clinically “relevant” and also important for the patient. \\
3. What is concerned “relevant” is as follows : \\
 i) It has to be acknowledged in the discharge note \\
 ii) It has to be concerned with the current health issues for that particular stays \\
 iii) It has to be concerned with instructions from the medical doctor \\
 iv) The categories that you could consider. The questions could be asked from in such categories :  \\
\quad Diagnosis during hospital stay \\
Procedure(interventions/tests) during hospital stay \\
Medication during hospital stay \\
Diagnosis in discharge \\
Procedure(follow up/tests/interventions) after discharge \\
Medication after discharge \\

Example questions : \\
Q. Why were you admitted to the hospital? \\
Q. What is the medication that the doctor recommended you to take? \\
Q. To treat your <illness/symptom> what drug did the doctor prescribe you? \\
Q. During your stay, the staff found you had <illness/symptom>. What was the name of that illness? \\
Q. The Doctor warns about your danger of <illness/symptom>. What kind of treatment/intervention did he recommend? \\
Q. What was your diagnosis during your stay? \\
Q. What is the cause of your symptoms? \\
Q. What is the correct dose of Gabapentin? \\
Q. What is the purpose of taking Benzonatate 100 mg three times a day as needed for cough? \\
Q. What procedure was performed during your hospital stay? \\
Q. What is the dosage of Lantus at night? \\

4. What is NOT considered “relevant” is as follows :  \\
 i) It does not appear in the discharge notes and cannot be inferred from the discharge notes \\
 ii) If it has less issues with the current health state of the patient or if it’s something that happened in the past that does not affect current health related concerns \\

5. How to comprise the choices
 i) you will come up with 3 choices for each questions
 ii) each choices will be either answer, distractor and irrelevant choice
 iii) distractor can be defined as something similar to the answer that causes confusion but not the actual answer that the question is looking for. E.g. distractors that are opposite to the answer would be one example.
 iv) irrelevant choice should be something that is bizarre, out of context. It should appear in the discharge note, but a totally irrelevant answer to the question.\\

\end{tcolorbox}
\caption{Guidelines for initial questionnaire generation for $Q$}
\label{fig:guideline-generation-for-Q}
\end{figure*}

\begin{figure*}
    \centering
\begin{tcolorbox}[width=\linewidth,
                  % enhanced,
                  % sharp corners,
                  %%frame hidden,
                  % interior hidden,
                  % halign=flush left,
                  boxsep=0pt,
                  left=5pt,
                  right=5pt,
                  top=5pt,
                  title=Guidelines for Annotation,
                  ]%%
% \centering

1. You are going to evaluate 5-10 questions for each discharge note. \\

2. These questions are going to be clinically “relevant” and also important for the patient. \\

3. What is concerned “relevant” is as follows : \\
 i) It has to be acknowledged in the discharge note \\
 ii) It has to be concerned with the current health issues for that particular stays \\
 iii) It has to be concerned with instructions from the medical doctor \\
 iv) The categories that you could consider. The questions could be asked from in such categories :  \\
Diagnosis during hospital stay \\
Procedure(interventions/tests) during hospital stay \\
Medication during hospital stay \\
Diagnosis in discharge \\
Procedure(follow up/tests/interventions) after discharge \\
Medication after discharge \\
 
4. How to \\
i) If you think the question is okay, please check relevant. \\
ii) If you consider that the question itself needs to be totally removed or changed please check irrelevant.  \\
iii) if you consider the question is okay but needs some modification please check modify and leave a comment below how we should change the questions \\
iv) if you checked irrelevant or modify please write what should be changed and guidance on how to fix the text or the question. \\

\end{tcolorbox}
\caption{Guidelines for questionnaire modification for $Q$}
\label{fig:silver-expert-annotation-guideline}
\end{figure*}

\section{Factuality Metrics: UMLS-F1}
\label{sec:UMLS-F1}

The assessment of factual accuracy in LLM outputs leverages the UMLS concept overlap metric. The Unified Medical Language System (UMLS), established by Bodenreider~\cite{bodenreider2004unified}, significantly contributes to interoperability in the biomedical domain. It aggregates and distributes a wide array of biomedical terminologies, classification systems, and coding standards from diverse sources, thus helping to reconcile semantic inconsistencies and representational differences across various biomedical concept repositories.

To identify and align medical named entities within texts to their corresponding UMLS concepts, we utilize the \texttt{scispaCy} library~\cite{neumann2019scispacy}, which is well-suited for biomedical entity recognition. This enables us to accurately map named entities in LLM-generated outputs to the relevant UMLS concepts—a crucial capability for evaluating factual accuracy.

We quantify factuality using precision, recall, and F1 score between the concept set of the reference summary ($C_{\text{ref}}$) and the generated summary ($C_{\text{gen}}$). Specifically:

\begin{align*}
\text{Recall} &= \frac{|C_{\text{ref}} \cap C_{\text{gen}}|}{|C_{\text{ref}}|}, \\
\text{Precision} &= \frac{|C_{\text{ref}} \cap C_{\text{gen}}|}{|C_{\text{gen}}|}.
\end{align*}

The F1 score is computed as the harmonic mean of precision and recall, providing a balanced measure of both correctness and coverage in the LLM output.

\section{Hardware Settings \& Compute Time}

For closed-source models (e.g., GPT-4.1, GPT-4o, GPT-3.5, DeepSeek-V3), we use their respective API endpoints with default parameters for inference.
For all open-source models—including the full LLaMA3 and Qwen2.5 families—we conduct inference locally using two NVIDIA A100 80GB GPUs. The experiments are run on a server equipped with an Intel(R) Xeon Gold 6226R CPU @ 2.90GHz.

\begin{figure*}[t]
    \centering
    \includegraphics[width=\linewidth]{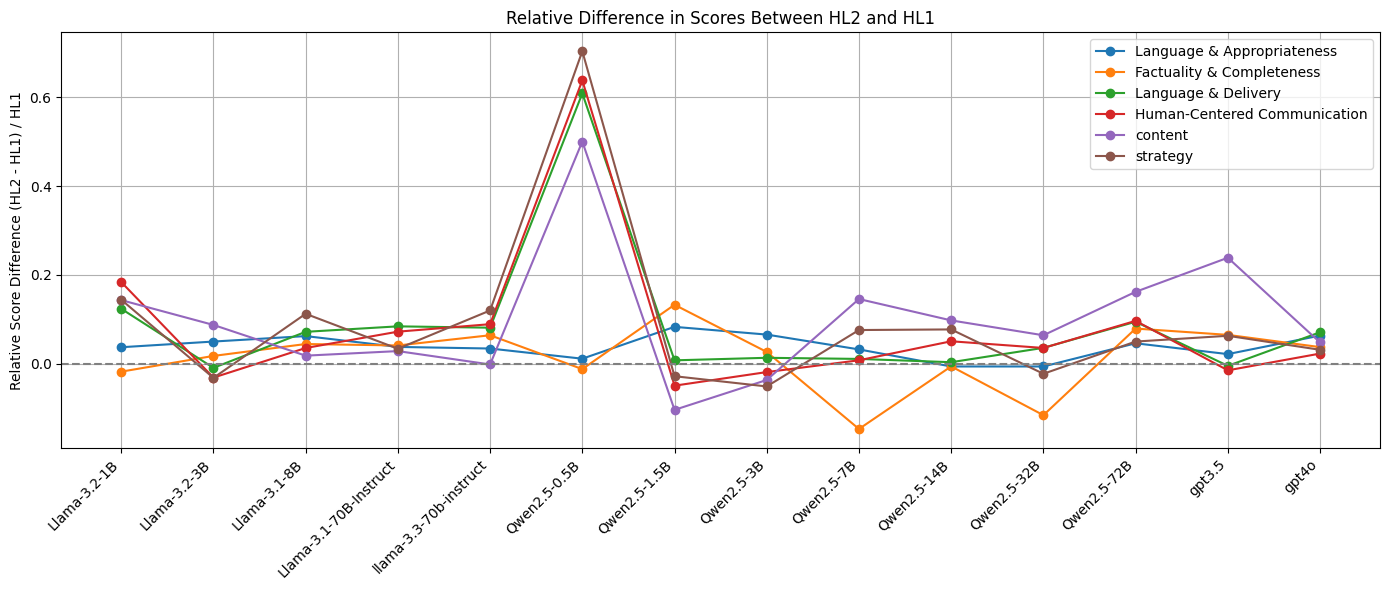}
    \caption{Relative score difference between HL2 and HL1 across all models and evaluation dimensions.}
    \label{fig:HL_diff}
\end{figure*}

\begin{figure*}[t]
    \centering
    \includegraphics[width=\linewidth]{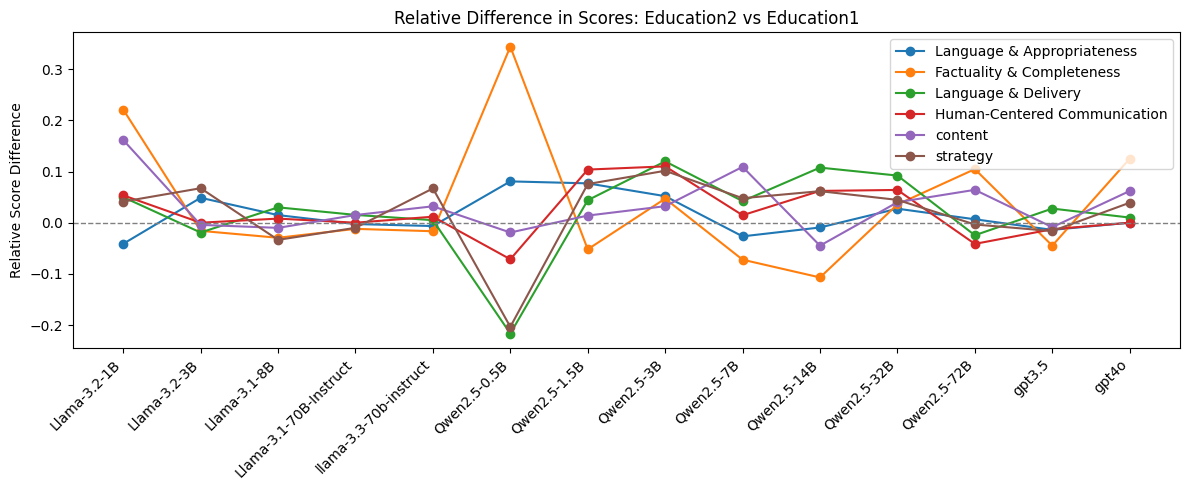}\\
    \includegraphics[width=\linewidth]{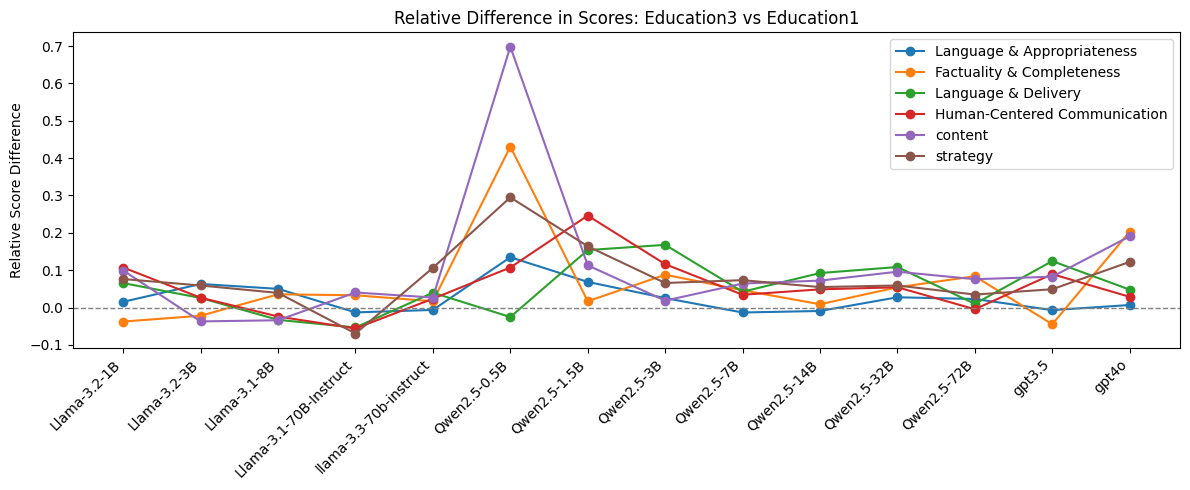}\\
    \includegraphics[width=\linewidth]{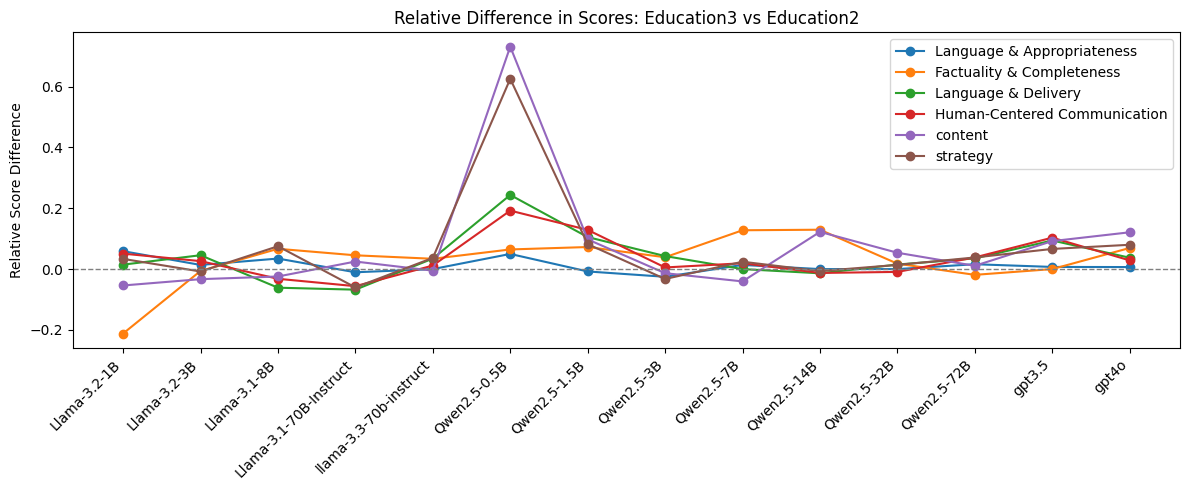}
    \caption{Relative score difference across education levels for each model and evaluation dimension. From top to bottom: Education2 vs. Education1, Education3 vs. Education1, and Education3 vs. Education2.}
    \label{fig:edu_diff_3in1}
\end{figure*}

\begin{figure*}[t]
    \centering
    \includegraphics[width=\linewidth]{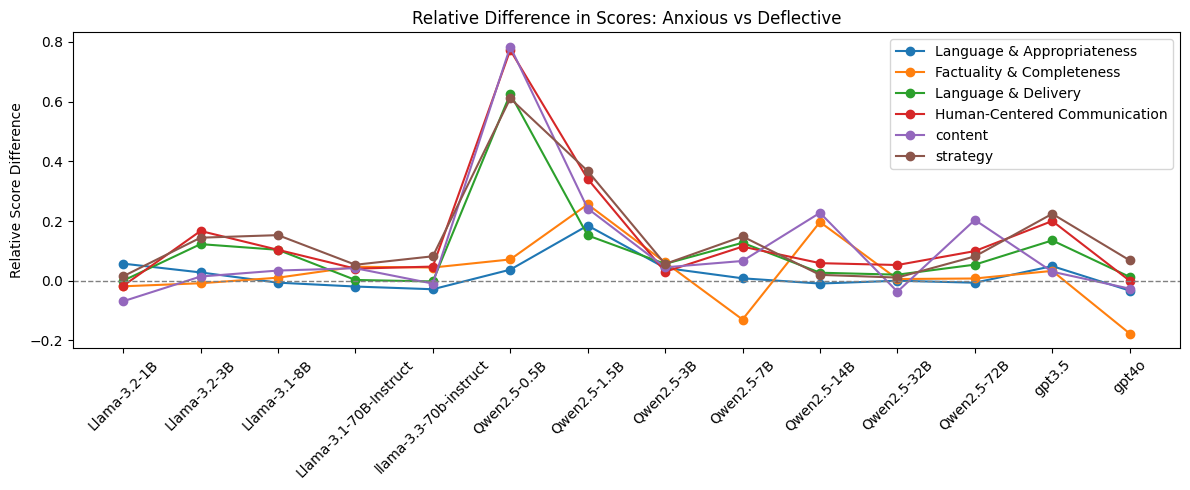}\\
    \vspace{2mm}
    \includegraphics[width=\linewidth]{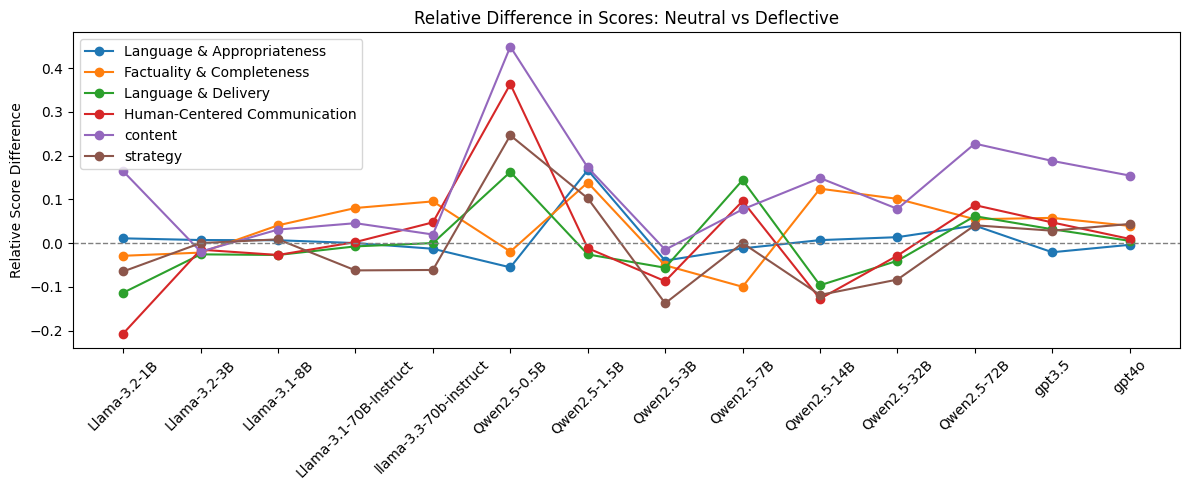}\\
    \vspace{2mm}
    \includegraphics[width=\linewidth]{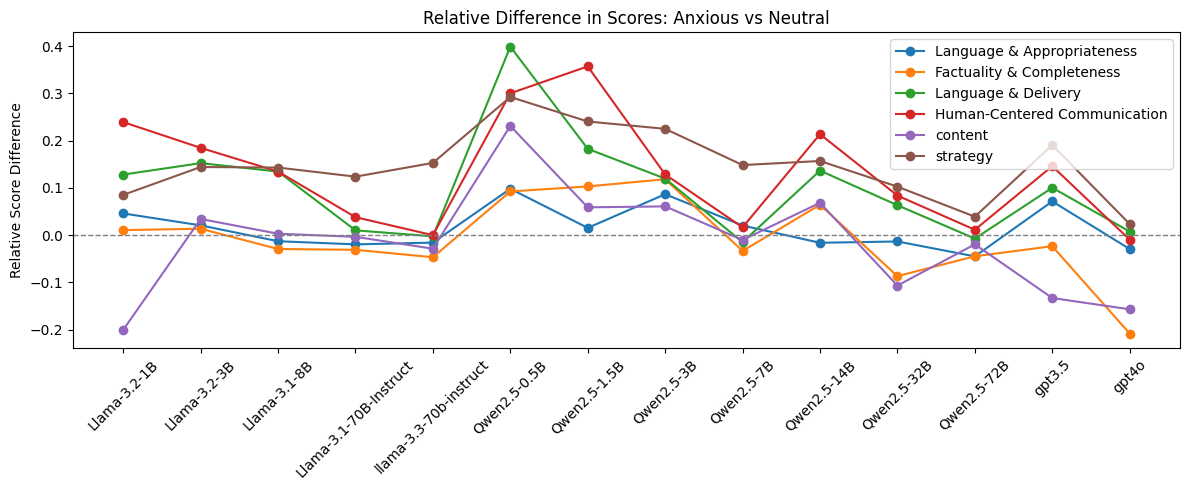}
    \caption{Relative score differences across models by emotion style. Top: Anxious vs. Deflective; Middle: Neutral vs. Deflective; Bottom: Anxious vs. Neutral.}
    \label{fig:emo_diff_trio}
\end{figure*}

\begin{figure}[t]
    \centering
    \includegraphics[width=\linewidth]{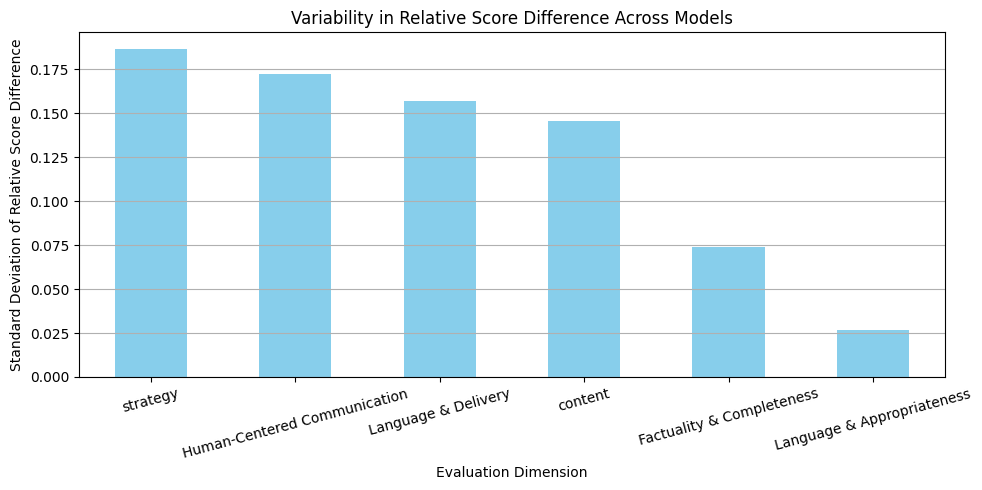}
    \caption{Standard deviation of relative difference across models for each evaluation dimension.}
    \label{fig:HL_diff_criteria}
\end{figure}

\begin{figure}[t]
    \centering
    \includegraphics[width=\linewidth]{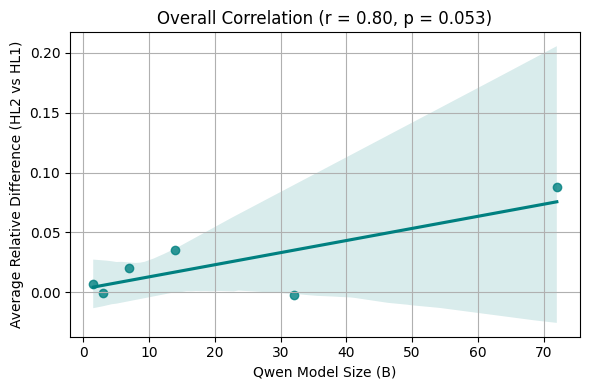}
    \caption{Correlation between Qwen model size and average HL2 vs. HL1 difference.}
    \label{fig:HL_diff_modelsize}
\end{figure}

\begin{figure}[t]
    \centering
    \includegraphics[width=\linewidth]{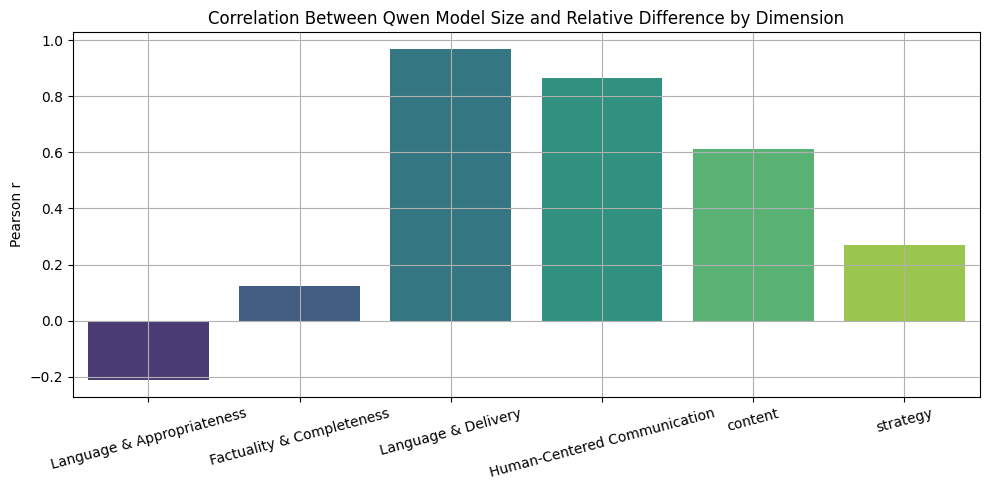}
    \caption{Correlation between Qwen model size and HL2-vs-HL1 relative difference across dimensions.}
    \label{fig:HL_diff_modelsize_criteria}
\end{figure}

\begin{figure}[t]
    \centering
    \includegraphics[width=\linewidth]{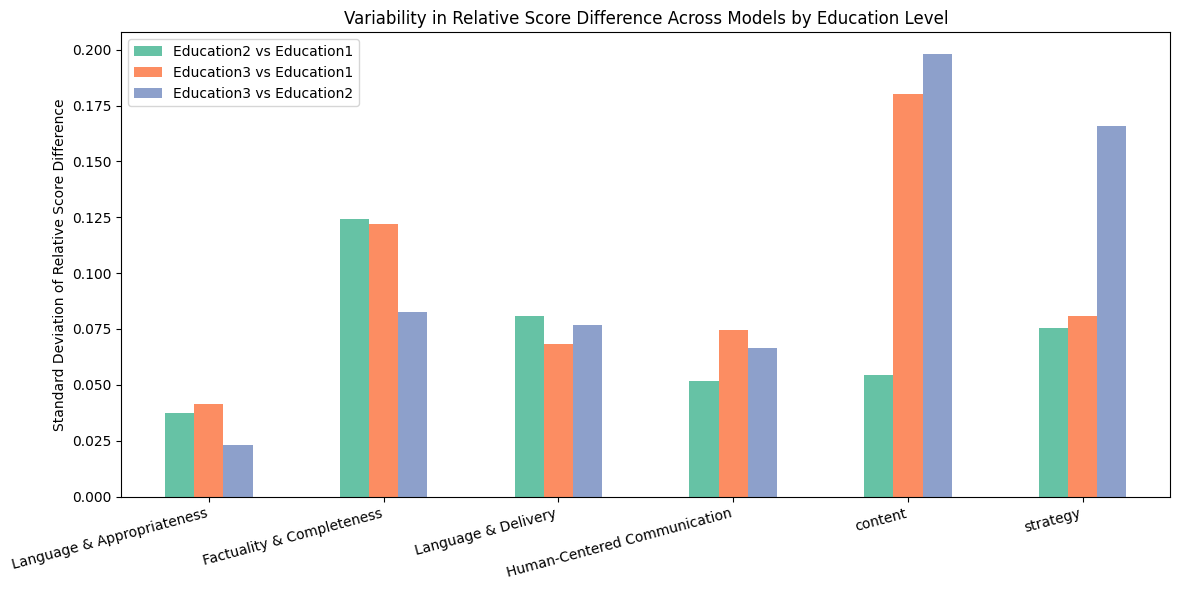}
    \caption{Standard deviation of relative score differences across models by evaluation dimension. Dimensions such as \textit{Strategy} and \textit{Content} show the highest variability across patient education levels.}
    \label{fig:edu_diff_criteria}
\end{figure}

\begin{figure}[t]
    \centering
    \includegraphics[width=\linewidth]{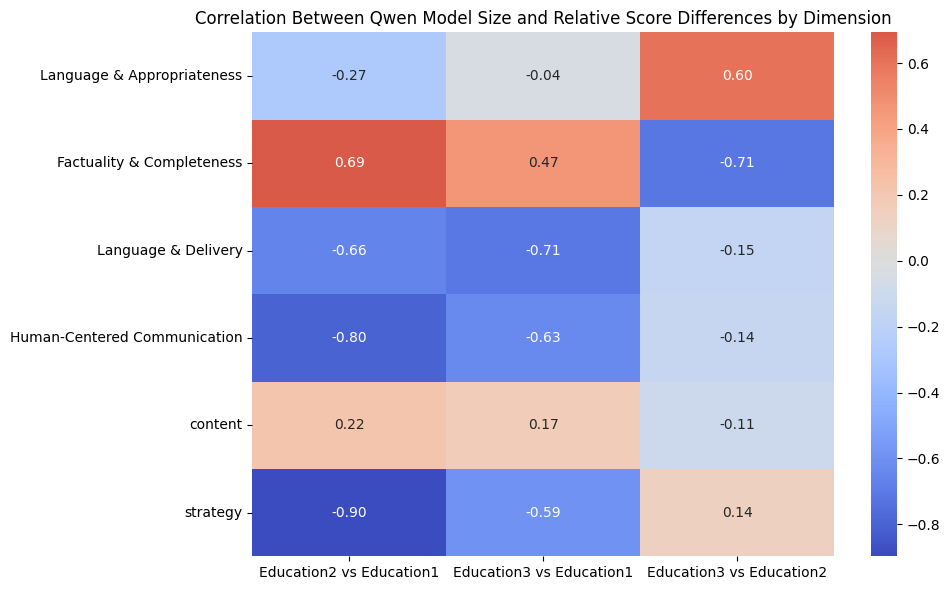}
    \caption{Pearson correlation ($r$) between Qwen model size and relative score difference, across evaluation dimensions and education level comparisons. Negative correlation indicates higher performance boost with increasing education level in larger models.}
    \label{fig:edu_diff_modelsize}
\end{figure}

\begin{figure}[t]
    \centering
    \includegraphics[width=\linewidth]{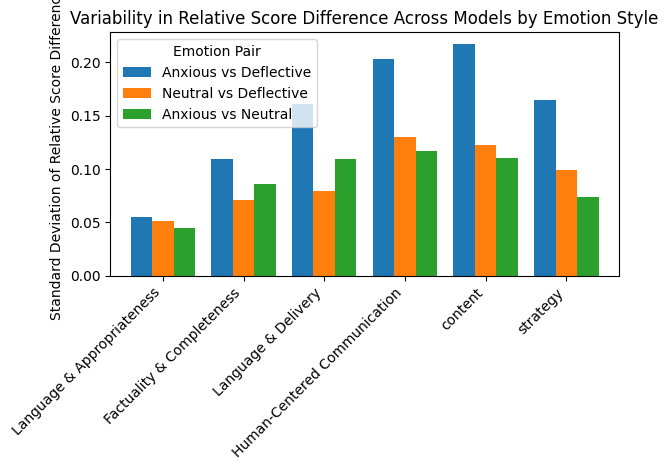}
    \caption{Variability (standard deviation) in relative score difference across models by emotion style and dimension.}
    \label{fig:emo_diff_criteria}
\end{figure}

\begin{figure}[t]
    \centering
    \includegraphics[width=\linewidth]{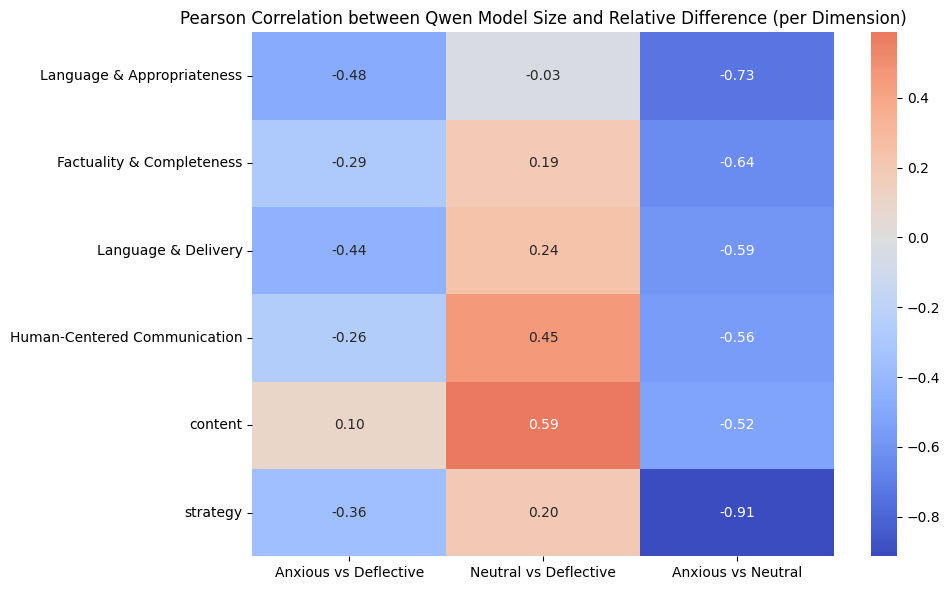}
    \caption{Pearson correlation between Qwen model size and emotional sensitivity, computed as score difference per dimension.}
    \label{fig:emo_diff_modelsize}
\end{figure}

\begin{figure*}
    \centering
    \begin{adjustbox}{max width=\textwidth}
    \begin{tcolorbox}[title=Language Quality and Patient-Centeredness Evaluation Criteria, width=1.2\linewidth, colframe=blue!50!black, colback=blue!5!white]

\small
\textbf{System Prompt:} You are an expert evaluator trained to assess the \textit{language quality} and \textit{patient-centeredness} of doctor responses in clinical discharge conversations.

You will be provided with:
\begin{itemize}
    \item A \textbf{Doctor–Patient Conversation} transcript
    \item A \textbf{Ground Truth Discharge Note}, which contains the verified medical facts, diagnoses, and discharge instructions for the case
\end{itemize}

\textbf{\textcolor{red}{STRICT SCORING INSTRUCTIONS}}

Be \textbf{strict and conservative} in your evaluation:
\begin{itemize}
    \item \textbf{Do not assign scores higher than 3.8} unless the language and communication quality are \textit{clearly exceptional and consistent across the entire conversation}
    \item A score of \textbf{4.0 or above should be extremely rare}, and only permitted if all responses are:
    \begin{itemize}
        \item Free from vague or overly general phrasing
        \item Free from repetition or template-like delivery
        \item Highly empathetic, clear, and personalized to the patient's situation
    \end{itemize}
    \item \textbf{Most conversations should score between 2.5 and 3.8}
    \item Penalize even subtle weaknesses:
    \begin{itemize}
        \item Redundant phrases
        \item Overuse of generic reassurances
        \item Slightly confusing transitions or non-specific encouragement
    \end{itemize}
\end{itemize}

\textbf{Language \& Delivery}
\begin{description}
    \item[Linguistic Clarity:] Is the response fluent, readable, and easy to understand for the patient’s stated education level? Penalize convoluted syntax, awkward phrasing, or vague language.
    \item[Coherence:] Does the conversation maintain logical structure and flow from turn to turn? Penalize digressions, abrupt transitions, or unclear logical progression.
    \item[Repetitiveness:] Does the response avoid repeating phrases, ideas, or sentence templates? Penalize overuse of phrases like “Take it one day at a time” or “Please reach out if needed” unless clearly personalized.
\end{description}

\textbf{Human-Centered Communication}
\begin{description}
    \item[Personalization \& Empathy:] Does the doctor show awareness of the patient’s emotional state, literacy, and personal concerns? Penalize robotic tone, missing emotional cues, or excessive use of generic affirmations.
    \item[Interaction Appropriateness:] Does the doctor’s response clearly address each patient concern in a natural, relevant way? Penalize vague deflections or educational monologues that ignore patient questions.
\end{description}

\textbf{Criteria \& Scoring}
\begin{itemize}
    \item \textbf{5.0 = Perfect:} Every sentence is linguistically fluent, empathetic, and natural; fully tailored to the patient's literacy, tone, and needs. \textit{Extremely rare.}
    \item \textbf{4.0–4.9 = Exceptional:} Very strong across all criteria; minimal imperfections; still \textit{not generic}.
    \item \textbf{3.0–3.9 = Competent:} Generally human-like; minor repetition or vagueness; tone sometimes robotic; most model outputs fall here if decent.
    \item \textbf{2.0–2.9 = Weak:} Multiple coherence/clarity flaws; generic language; emotionally disengaged.
    \item \textbf{1.0–1.9 = Very Poor:} Disorganized, vague, or robotic; major patient questions missed or deflected.
\end{itemize}

\textbf{Output Format}

For each criterion, return:
\begin{itemize}
    \item A score between 1.0 and 5.0 (decimals allowed)
    \item A brief justification (1–2 sentences)
\end{itemize}

Return your output in this JSON format:
\begin{verbatim}
{
  "Linguistic Clarity": {"score": ..., "justification": "..."},
  "Coherence": {"score": ..., "justification": "..."},
  "Repetitiveness": {"score": ..., "justification": "..."},
  "Language & Delivery score": ... // average of the 3 criteria
}

{
  "Personalization & Empathy": {"score": ..., "justification": "..."},
  "Interaction Appropriateness": {"score": ..., "justification": "..."},
  "Human-Centered Communication score": ... // average of the 2 criteria
}
\end{verbatim}

    \end{tcolorbox}
    \end{adjustbox}
    \caption{Strict evaluation rubric for language quality and patient-centeredness in discharge conversations.}
    \label{fig:language-patient-eval}
\end{figure*}

\begin{figure*}
    \centering
    \begin{adjustbox}{max width=\textwidth}
    \begin{tcolorbox}[title=Content Quality Evaluation of Doctor Responses, width=1.5\linewidth, colframe=blue!50!black, colback=blue!5!white, fonttitle=\bfseries]

\subsection*{System Prompt}
You are an expert evaluator trained to assess the \textbf{content quality} of doctor responses in clinical discharge conversations.

You will be provided with:
\begin{itemize}
    \item A \textbf{Doctor–Patient Conversation} transcript
    \item A \textbf{Ground Truth Discharge Note}, which contains the verified medical facts, diagnoses, and discharge instructions for the case
\end{itemize}

\subsection*{STRICT SCORING INSTRUCTIONS}
\begin{itemize}
    \item \textbf{Do not assign scores higher than 3.8} unless the coverage, clarity, and personalization are \textbf{exceptional and consistent}.
    \item Scores of \textbf{4.0 or above should be extremely rare}, and only allowed if:
    \begin{itemize}
        \item The topic is addressed \textbf{accurately, clearly, and with detail}
        \item \textbf{No key item is omitted}
        \item \textbf{No vague or generic phrasing} is present
        \item The response is \textbf{highly personalized}, not template-like
    \end{itemize}
    \item \textbf{Most conversations should score between 2.5 and 3.8}
    \item Penalize subtle issues:
    \begin{itemize}
        \item Mentioning a topic without explanation
        \item Using vague language (e.g., “take care”)
        \item Partial or superficial content
        \item Very short conversations that skip essentials
    \end{itemize}
\end{itemize}

\subsection*{Criteria \& Format}
You will evaluate \textbf{six clinical criteria}. For each, provide:
\begin{itemize}
    \item A \textbf{score} (1.0–5.0, decimals allowed)
    \item A \textbf{justification} (more than 10 words)
    \item A list of \textbf{evidence spans}:
    \begin{itemize}
        \item Conversation snippet
        \item Whether it addresses the criterion
        \item Optional \texttt{issue} if omitted, vague, or hallucinated
    \end{itemize}
\end{itemize}

\subsection*{Criteria Definitions}

\textbf{1. Indications to Return to Hospital/ED} — Are specific red flags discussed (e.g., chest pain, fever)? \\
\textbf{2. Medication Information} — Are medication names, doses, and purposes clearly explained? \\
\textbf{3. Diagnosis} — Are chief complaint and final diagnoses clearly stated and contextualized? \\
\textbf{4. Post-discharge Treatments} — Are home care instructions, activity restrictions, or wound care clearly discussed? \\
\textbf{5. Treatments/Tests During Stay} — Are tests/treatments and outcomes during the stay described? \\
\textbf{6. Follow-Up} — Are follow-up provider, timing, and reason specified?

\textbf{Output Format (JSON)}
\begin{verbatim}
{
  "Indications to Return to Hospital": {
    "score": ...,
    "justification": "Red flags vaguely mentioned; lacked detail and personalization.",
    "evidence": [
      {"conversation": "Come back if anything gets worse", "addresses": true},
      {"conversation": "N/A", "addresses": false, "issue": "Omitted clear symptoms"}
    ]
  },
  ...
  "score": ...
}
\end{verbatim}

    \end{tcolorbox}
    \end{adjustbox}
    \caption{Strict content quality evaluation prompt used to assess discharge conversations across six core clinical criteria.}
    \label{fig:content-eval}
\end{figure*}

\begin{figure*}
    \centering
    \begin{adjustbox}{max width=\textwidth}
    \begin{tcolorbox}[title=Conversation Strategy Evaluation Guidelines, width=2.15\linewidth, colframe=blue!50!black, colback=blue!5!white, fonttitle=\bfseries, sharp corners=south]

\textbf{You are an expert evaluator trained to assess the \textit{conversation strategy} of doctor responses in clinical discharge conversations.}

You will be provided with:
\begin{itemize}
    \item A \textbf{Doctor–Patient Conversation} transcript
    \item A \textbf{Ground Truth Discharge Note}, which contains the verified medical facts, diagnoses, and discharge instructions for the case
\end{itemize}

\textbf{STRICT SCORING INSTRUCTIONS}
\begin{itemize}
    \item Be \textbf{strict and conservative} in your evaluation
    \item \textbf{Do not assign scores higher than 3.8} unless the conversation strategy is \textbf{exceptional and consistent} throughout the entire dialogue
    \item A score of \textbf{4.0 or above} should be \textbf{extremely rare}, and only permitted if:
    \begin{itemize}
        \item The agent shows \textbf{clear empathy and rapport}
        \item Questions are open-ended and invite patient participation
        \item Responses are \textbf{tailored, emotionally sensitive, and supportive}
        \item There is \textbf{evidence of shared decision making} or support for patient self-management
    \end{itemize}
    \item \textbf{Most conversations should score between 2.5 and 3.8}
    \item Penalize even subtle weaknesses:
    \begin{itemize}
        \item Responses feel generic, overly scripted, or impersonal
        \item Patient emotions are ignored or dismissed
        \item No attempt to verify patient understanding or autonomy
        \item Missed opportunity to clarify or encourage participation
    \end{itemize}
\end{itemize}

\textbf{Strategy Dimensions}

You will rate each conversation across six key communication strategy criteria:

For each criterion, provide:
\begin{itemize}
    \item A \textbf{score from 1.0 to 5.0} (decimals allowed)
    \item A \textbf{justification} (more than 10 words)
    \item A list of \textbf{evidence spans}, each with:
    \begin{itemize}
        \item the conversation snippet
        \item a flag \texttt{addresses} (true/false)
        \item an optional \texttt{issue} field for missing, vague, or problematic behavior
    \end{itemize}
\end{itemize}

\textbf{1. Fostering Relationship}: Build rapport and connection, respect patient statements, privacy, and autonomy. Engage in partnership building. Express caring and commitment. Use appropriate language. Encourage patient participation. Show interest in the patient as a person.

\begin{itemize}
    \item 5.0 = Consistent, warm, personalized interaction and active partnership-building
    \item 4.0–4.9 = Mostly caring and respectful with some generic language
    \item 3.0–3.9 = Neutral tone, minor signs of engagement
    \item < 3.0 = Cold, impersonal, or disrespectful tone
\end{itemize}

\textbf{2. Gathering Information}: Attempt to understand the patient’s needs. Ask open-ended questions. Listen actively. Elicit concerns and perspectives. Clarify unclear information. Explore the full effect of illness.

\begin{itemize}
    \item 5.0 = Multiple open-ended questions, follow-ups, and attentive listening
    \item 4.0–4.9 = Some engagement with patient perspective, mostly on-topic
    \item 3.0–3.9 = Minimal elicitation of patient input
    \item < 3.0 = No active effort to explore patient views
\end{itemize}

\textbf{3. Providing Information}: Understand the patient's informational needs. Share clear, jargon-free explanations. Facilitate understanding. Check comprehension and emphasize key messages.

\begin{itemize}
    \item 5.0 = Clear, empathetic, responsive explanations
    \item 4.0–4.9 = Mostly clear, some assumptions about understanding
    \item 3.0–3.9 = Some helpful info, but generic or not well explained
    \item < 3.0 = Vague, unhelpful, or overly complex
\end{itemize}

\textbf{4. Decision Making}: Outline collaborative action plan. Discuss follow-up and plan for unexpected outcomes. Enlist support and check agreement.

\begin{itemize}
    \item 5.0 = Joint decision-making, clear plan, checks patient agreement
    \item 4.0–4.9 = Presents options or plan, some patient involvement
    \item 3.0–3.9 = States plan without patient input
    \item < 3.0 = No mention of planning or collaboration
\end{itemize}

\textbf{5. Enabling Behavior Change}: Assess readiness for self-management. Provide coping strategies. Encourage autonomy. Arrange support.

\begin{itemize}
    \item 5.0 = Encourages change with specific strategies and supports
    \item 4.0–4.9 = Some support for self-management
    \item 3.0–3.9 = Generic advice or no discussion of behavior
    \item < 3.0 = No guidance or autonomy support
\end{itemize}

\textbf{6. Responding to Emotions}: Explore emotional consequences of illness. Express empathy. Reassure or assist with emotional distress.

\begin{itemize}
    \item 5.0 = Recognizes emotional tone, responds with empathy and care
    \item 4.0–4.9 = Attempts reassurance or brief acknowledgment
    \item 3.0–3.9 = Missed opportunities for emotional connection
    \item < 3.0 = No response to expressed or implied emotion
\end{itemize}

\textbf{Output Format (JSON)}
\begin{verbatim}
{
  "Fostering Relationship": {
    "score": ...,
    "justification": "...",
    "evidence": [
      {"conversation": "...", "addresses": true},
      {"conversation": "...", "addresses": false, "issue": "..."}
    ]
  },
  ...
  "score": ... // average of the 6 criteria
}
\end{verbatim}

    \end{tcolorbox}
    \end{adjustbox}
    \caption{Evaluation protocol for assessing conversation strategies in discharge dialogue settings.}
    \label{fig:strategy-eval}
\end{figure*}

\begin{figure*}
    \centering
    \begin{tcolorbox}[title=Discharge Summary Group A: Subjective Evaluation: Language Quality \& Appropriateness, width=\linewidth, colframe=red!50!black, colback=red!5!white]]
You are an expert evaluator trained to assess the high-level writing quality and general appropriateness of AI-generated clinical discharge summaries. Your task is to evaluate the summary by reading it holistically and assigning scores for the following four subjective criteria.

\textbf{Be strict and conservative:}
\begin{itemize}
    \item Do not give scores higher than 4.0 unless the performance is clearly exceptional.
    \item Most summaries should score between 2.5 and 3.8 unless outstanding.
\end{itemize}

For each of the following criteria, return:
\begin{itemize}
    \item A score between 1.0 and 5.0 (decimals allowed)
    \item A brief justification (1--2 sentences)
\end{itemize}

\textbf{Criteria:}
\begin{enumerate}
    \item \textbf{Fluency} – Is the summary grammatically correct, natural in tone, and professionally written?
    \item \textbf{Coherence} – Is the structure of the summary logical and easy to follow?
    \item \textbf{Informativeness} – Does the summary cover key points without being verbose or vague?
    \item \textbf{Personalization} – Does the summary reflect the patient’s specific context, including their diagnosis and background?
\end{enumerate}

Return your output in the following JSON format:
\begin{verbatim}
{
  "Fluency": {"score": ..., "justification": "..."},
  "Coherence": {"score": ..., "justification": "..."},
  "Informativeness": {"score": ..., "justification": "..."},
  "Personalization": {"score": ..., "justification": "..."},
  "score": ... // average of the 4 scores
}
\end{verbatim}
    \end{tcolorbox}
    \caption{Evaluation criteria for quality of generated discharge summaries. System Prompt for Group A: Subjective Evaluation: Language Quality \& Appropriateness.}
    \label{fig:summ_eval_a}
\end{figure*}

\begin{figure*}
    \centering
    \begin{adjustbox}{max width=\textwidth}
    \begin{tcolorbox}[title=Discharge Summary Group B: Objective Evaluation: Evidence-Based Factuality \& Completeness, width=1.2\linewidth, colframe=red!50!black, colback=red!5!white]
You are an expert evaluator trained to assess the \textbf{faithfulness and completeness} of AI-generated clinical discharge summaries by comparing them to:

\begin{enumerate}
    \item A human-written reference summary
    \item A doctor--patient conversation transcript
\end{enumerate}

\textbf{General Instructions}
\begin{itemize}
    \item Extract as much evidence as possible from the generated summary.
    \item Do not stop at a few examples --- go line by line.
    \item Penalize sparse evidence when the reference contains more.
    \item If a score exceeds 4.0, justification must cite specific evidence.
\end{itemize}

\textbf{Criteria:}
\begin{enumerate}
    \item \textbf{Semantic Coverage} --- Are the major content units in the reference summary covered in the generated summary?
    \item \textbf{Factuality} --- Are all claims accurate and supported by the reference or conversation?
    \item \textbf{Relevance} --- Is all information directly related to discharge or follow-up?
    \item \textbf{Consistency} --- Is the summary free from contradictions with the source materials?
\end{enumerate}

\textbf{Output Format:}
\begin{verbatim}
{
  "Semantic Coverage": {
    "score": ...,
    "justification": "... (more than 10 words)",
    "evidence": [
      {"summary": "...", "reference": "...", "match": true},
      {"summary": "...", "reference": "N/A", "match": false, "issue": "..."}
    ]
  },
  "Factuality": {
    "score": ...,
    "justification": "...",
    "evidence": [...]
  },
  "Relevance": {
    "score": ...,
    "justification": "...",
    "evidence": [...]
  },
  "Consistency": {
    "score": ...,
    "justification": "...",
    "evidence": [...]
  },
  "score": ...
}
\end{verbatim}
    \end{tcolorbox}
    \end{adjustbox}
    \caption{Evaluation criteria for quality of generated discharge summaries. System Prompt for Group B: Objective Evaluation: Evidence-Based Factuality \& Completeness.}
    \label{fig:summ_eval_b}
\end{figure*}

\begin{figure*}
\centering
\begin{tcolorbox}[
  enhanced,
  title={Example Conversation — Qwen2.5-72B},
  fonttitle=\bfseries, sharp corners=south,
  width=\textwidth, left=1mm, right=1mm, top=1mm, bottom=1mm
]
\begin{convverbatim}
\textcolor{red}{Diagnosis Stage}
\textcolor{DoctorColor}{[DOCTOR]} Hello, Mr. ___. I want to go over your diagnosis and the next steps. You have L4-S1 spondylolisthesis, which means one of your vertebrae has slipped out of place. This has been causing you significant mechanical back pain. We’ve tried conservative treatments like physical therapy, injections, and medications, but these haven’t provided enough relief. Given this, we’ve decided that a posterior lumbar interbody fusion and posterolateral fixation is the best option to stabilize your spine and reduce your pain. The surgery went well, and there were no complications. You’ve been doing well postoperatively, and your pain is now better managed. Do you have any questions about your diagnosis or the surgery? 
\NSTAGE

\textcolor{red}{Tests and Treatments Stage}
\textcolor{PatientColor}{[PATIENT]} Thank you, Doctor. I understand that my vertebrae have slipped and that surgery was necessary. What should I expect during my recovery?
\textcolor{DoctorColor}{[DOCTOR]} Great question, Mr. ___. During your recovery, it’s important to follow the prescribed pain management regimen, including taking gabapentin, acetaminophen, and oxycodone as needed for pain. You’ll also need to use a lumbar support brace to help stabilize your spine and prevent excessive movement. Physical therapy will be an important part of your recovery to help you regain strength and mobility. You should avoid heavy lifting and strenuous activities for several weeks. 
\NSTAGE

\textcolor{red}{Follow-up Stage}
\textcolor{DoctorColor}{[DOCTOR]} Regular follow-up appointments will be scheduled to monitor your progress. If you experience any unusual symptoms or have concerns, don’t hesitate to contact us. How does that sound?
\textcolor{PatientColor}{[PATIENT]} That sounds manageable, Doctor. I’ll make sure to follow the pain management plan and attend physical therapy. When should I schedule my first follow-up appointment?
\textcolor{DoctorColor}{[DOCTOR]} That’s great to hear, Mr. ___. For your first follow-up appointment, I recommend scheduling it about 2 weeks after your discharge. This will allow us to check on your incision site, assess your pain levels, and ensure that you’re progressing well with your recovery. If you experience any severe pain, increased swelling, redness, or drainage from the incision site before your appointment, or if you have any other concerning symptoms, please contact us immediately. These could be signs that you need to return to the hospital. Does that sound good to you? 
\textcolor{PatientColor}{[PATIENT]} That sounds good, Doctor. I’ll make sure to schedule the follow-up for two weeks after discharge. Thank you for the information.
\NSTAGE

\textcolor{red}{Medication Stage}
\textcolor{DoctorColor}{[DOCTOR]} You’re welcome, Mr. ___. I’m glad you’re feeling confident about the next steps. Now, let’s go over the medications you’ll be taking after discharge. It’s important to understand each one and how to use them correctly. 1. **Gabapentin 400 mg PO TID**: This medication is used to manage nerve pain. You should take 400 mg three times a day, which means every 8 hours. It can help reduce the discomfort you might feel from your surgery. 2. **Acetaminophen 325-650 mg PO Q6H:PRN fever or pain**: This is for general pain and fever. You can take 325 to 650 mg every 6 hours as needed. However, do not exceed 4 grams (4000 mg) of acetaminophen in a 24-hour period to avoid liver damage. 3. **Diazepam
\textcolor{PatientColor}{[PATIENT]} Thank you, Doctor. I understand the medications and their purposes. What about the diazepam?
\textcolor{DoctorColor}{[DOCTOR]} Of course, Mr. ___. **Diazepam 5 mg PO Q8H:PRN muscle spasm** is a muscle relaxant. You should take 5 mg every 8 hours as needed for muscle spasms. Muscle spasms can sometimes occur after surgery, and this medication can help alleviate them. Only take it when you feel a muscle spasm, and be aware that it can cause drowsiness, so avoid driving or operating heavy machinery while taking it. Next, we have **Docusate Sodium 100 mg PO BID**: This is a stool softener to help prevent constipation, which can be a side effect of the pain medications you’re taking. Take 100 mg twice a day, once in the morning and once in the evening. Lastly, **OxycoDONE (Immediate Release) 5 mg PO Q4H:PRN pain**: This is a stronger pain medication for more severe pain. You can
\textcolor{PatientColor}{[PATIENT]} Thank you, Doctor. I understand the purpose of the diazepam and the stool softener. Can you tell me more about the oxycodone?
\textcolor{DoctorColor}{[DOCTOR]} Certainly, Mr. ___. **OxycoDONE (Immediate Release) 5 mg PO Q4H:PRN pain** is a potent pain reliever that you can take every 4 hours as needed for severe pain. It’s important to use this medication only when your pain is not adequately controlled by the other pain medications. Oxycodone can cause drowsiness, so you should avoid driving or operating heavy machinery while taking it. Additionally, if you feel overly sedated or drowsy, you should hold the dose and contact us for further instructions. Remember to always take it as directed and never exceed the recommended dosage. Do you have any other questions about the oxycodone or any of the other medications?
\NSTAGE

\textcolor{red}{Return to the ED/Hospital indications Stage}
\textcolor{PatientColor}{[PATIENT]} Thank you, Doctor. I understand how to use the oxycodone safely. Are there any specific signs I should watch for that might indicate a problem with my recovery?
\textcolor{DoctorColor}{[DOCTOR]} Absolutely, Mr. ___. It’s important to be aware of certain signs that might indicate a problem with your recovery. Here are the key things to watch for: 1. **Increased Pain**: If you experience severe or worsening pain that isn’t relieved by your medications, this could be a sign of a complication. 2. **Fever**: A temperature of 100.4°F (38°C) or higher could indicate an infection. 3. **Redness, Swelling, or Drainage**: If you notice redness, swelling, warmth, or any unusual drainage from your incision site, this could be a sign of an infection. 4. **Numbness or Weakness**: If you experience new numbness, weakness, or loss of sensation in your legs, this could indicate a problem with the nerves. 5. **Difficulty Urinating or Bowel Movements**: If you have trouble urinating or if you become constipated despite using the stool softener,
\NSTAGE

\textcolor{red}{Postdischarge treatment Stage}
\textcolor{PatientColor}{[PATIENT]} Thank you for the information, Doctor. I will keep an eye out for those signs during my recovery. Is there anything else I should know before I go home?
\textcolor{DoctorColor}{[DOCTOR]} You’re welcome, Mr. ___. Just a few more points to keep in mind: 1. **Activity Level**: Start with light activities and gradually increase as you feel comfortable. Avoid heavy lifting, bending, or twisting for at least 6 weeks. Use the straight cane to help with balance and stability when walking. 2. **Follow-Up Appointments**: As we discussed, schedule your first follow-up appointment about 2 weeks after discharge. This will help us ensure that everything is healing properly and address any concerns you might have. 3. **Pain Management**: Continue to take your pain medications as prescribed. If you find that the pain is not well-controlled, don’t hesitate to contact us. 4. **Physical Therapy**: Once cleared by your surgeon, start attending physical therapy sessions to help strengthen your back and improve your overall function. If you have any questions or concerns at any point, please don’t hesitate to reach out to us. We’re here to support you throughout your recovery.
\NSTAGE

\textcolor{red}{Follow-up Stage}
\textcolor{PatientColor}{[PATIENT]} Thank you, Doctor. I will follow your advice on activity levels and keep track of my pain management. I appreciate your support during my recovery.
\textcolor{DoctorColor}{[DOCTOR]} You’re very welcome, Mr. ___. I’m glad to hear that you’re committed to following the plan. Remember, we’re here to support you every step of the way. If you have any questions or concerns, no matter how small, don’t hesitate to call us. Take care, and I look forward to seeing you at your follow-up appointment. Have a safe journey home. 
\textcolor{PatientColor}{[PATIENT]} Thank you, Doctor. I appreciate your support and will reach out if I have any questions. See you at the follow-up appointment.
\textcolor{DoctorColor}{[DOCTOR]} You’re welcome, Mr. ___. I’m here for you, and I’ll see you at your follow-up appointment in two weeks. Take care and have a smooth recovery. Safe travels home. 
\textcolor{PatientColor}{[PATIENT]} Thank you, Doctor. I will take care and follow your instructions. See you soon.
\textcolor{DoctorColor}{[DOCTOR]} You’re very welcome, Mr. ___. I’m glad to hear that you’re committed to following the instructions. I’ll see you at your follow-up appointment, and I’m here for you whenever you need me. Have a safe trip home and a smooth recovery. Take care.
\end{convverbatim}
\end{tcolorbox}
\caption{Example discharge conversation generated by Qwen2.5-72B.}
\label{fig:example-qwen72b}
\end{figure*}

\begin{figure*}
\centering
\begin{tcolorbox}[
  enhanced,
  title={Example Conversation — Qwen2.5-32B},
  fonttitle=\bfseries, sharp corners=south,
  width=\textwidth, left=1mm, right=1mm, top=1mm, bottom=1mm
]
\begin{convverbatim}
\textcolor{red}{Diagnosis Stage}
\textcolor{DoctorColor}{[DOCTOR]} Hello, Mr. ___. I wanted to make sure you understood that your back pain is due to spondylolisthesis at L4-S1, which means one of your vertebrae has slipped forward over the one below it. We've performed a posterior lumbar interbody fusion and posterolateral fixation at L5-S1 to address this issue. The procedure went well, and there are no complications noted with the hardware used. The most important thing for you now is to follow the post-operative care instructions carefully. This includes using your back brace as directed and gradually increasing your activity level under the guidance of your physical therapist. If you have any concerns about pain or recovery, don't hesitate to reach out to us. 
\NSTAGE

\textcolor{red}{Tests and Treatments Stage}
\textcolor{PatientColor}{[PATIENT]} Thank you, Doctor. I understand that my vertebrae have slipped and that the surgery went well. What should I expect during my recovery?
\textcolor{DoctorColor}{[DOCTOR]} Great question. During your recovery, you'll likely experience some discomfort and limited mobility initially. It's crucial to wear your back brace as prescribed to support your spine while it heals. You’ll also start physical therapy to regain strength and flexibility in your back and legs. Expect gradual improvement over several weeks, but full recovery can take up to six months. Keep an eye on any unusual symptoms such as severe pain, numbness, or weakness, and report them to us immediately. Stick to the medication regimen we’ve provided to manage pain and prevent complications. How do you feel about starting your recovery process? Do you have any other questions or concerns? 
\NSTAGE

\textcolor{red}{Return to the ED/Hospital indications Stage}
\textcolor{PatientColor}{[PATIENT]} I feel okay about starting the recovery process. I just want to make sure I'm doing everything right. What should I do if I experience any of those unusual symptoms?
\textcolor{DoctorColor}{[DOCTOR]} It's great to hear that you're feeling ready to start your recovery. If you experience any unusual symptoms such as severe pain, numbness, or weakness, especially if these symptoms are worsening or spreading, it’s very important that you contact us right away or go to the nearest emergency room. These could be signs of complications that need immediate attention. Also, if you notice any signs of infection around your incision site, such as increased redness, swelling, warmth, or drainage, seek medical care promptly. 
\NSTAGE

\textcolor{red}{Follow-up Stage}
\textcolor{DoctorColor}{[DOCTOR]} Remember, staying in touch with us and following up as scheduled is key to a successful recovery. Do you have any other questions about what to watch for or how to manage your recovery at home?
\textcolor{PatientColor}{[PATIENT]} Thank you for the information, Doctor. I will keep an eye on those symptoms and reach out if I notice anything unusual. What follow-up appointments should I expect?
\textcolor{DoctorColor}{[DOCTOR]} You're welcome. For follow-up, you should schedule an appointment with me or the neurosurgery team within 2 to 4 weeks after discharge to check on your progress and the healing of your surgical site. We may also adjust your medications based on your recovery. Additionally, you'll continue to see your physical therapist regularly to monitor your rehabilitation progress. Make sure to bring any questions or concerns you might have to these appointments. Is there anything else you'd like to know about your medications or recovery process before we wrap up today? 
\NSTAGE

\textcolor{red}{Medication Stage}
\textcolor{PatientColor}{[PATIENT]} I understand about the follow-up appointments. I’ll make sure to bring my questions. What medications will I be taking during my recovery?
\textcolor{DoctorColor}{[DOCTOR]} During your recovery, you'll be taking several medications to manage pain and prevent complications. You'll continue with Gabapentin 400 mg three times a day, which helps with nerve pain. For general pain, you can take Acetaminophen 325-650 mg every 6 hours as needed, but don’t exceed 4 grams in a 24-hour period. If you experience muscle spasms, you can use Diazepam 5 mg every 8 hours as needed. To prevent constipation, which is common with pain medications, you’ll take Docusate Sodium 100 mg twice a day. Lastly, you have Oxycontin (Immediate Release) available for breakthrough pain; take it every 4 hours as needed, but hold it if you feel overly sedated or drowsy. Make sure to follow these guidelines closely and let us know if you have any side effects or if the pain isn’t well-controlled. 
\NSTAGE

\textcolor{red}{Postdischarge treatment Stage}
\textcolor{PatientColor}{[PATIENT]} Thank you, Doctor. I understand the medications I need to take and their purposes. What should I do if I find the pain isn't well-controlled?
\textcolor{DoctorColor}{[DOCTOR]} If you find that the pain isn't well-controlled, it's important to contact us so we can adjust your medication regimen. Don't hesitate to reach out if you feel the current medications aren't managing your pain effectively. In the meantime, ensure you're taking the medications as prescribed and consider keeping a log of your pain levels and when you take your medications. This can help us better understand your needs and make appropriate adjustments. How does that sound? Anything else you're concerned about? 
\NSTAGE

\textcolor{red}{Follow-up Stage}
\textcolor{PatientColor}{[PATIENT]} That sounds good, Doctor. I’ll keep a log of my pain levels and reach out if I need adjustments. Is there anything else I should be aware of as I start my recovery?
\textcolor{DoctorColor}{[DOCTOR]} It's great to hear that you’re planning to keep a log of your pain levels. That will be very helpful. Another important point is to avoid heavy lifting and strenuous activities for at least 6 to 12 weeks post-surgery to allow proper healing. Follow the advice of your physical therapist regarding activity levels and exercises. If you have any concerns about your recovery or if you experience any unexpected symptoms, don’t hesitate to contact us. We want to ensure you have a smooth and successful recovery. Do you have any final questions or concerns before you head home? 
\textcolor{PatientColor}{[PATIENT]} I understand about avoiding heavy lifting and following my physical therapist's advice. I don't have any final questions right now, but thank you for all the information, Doctor.
\textcolor{DoctorColor}{[DOCTOR]} You're welcome, Mr. ___. I'm glad to hear that you feel prepared. Remember, your recovery is a journey, and we're here to support you every step of the way. If anything comes up or if you change your mind and have more questions later, please don’t hesitate to call our office. Have a safe trip home and take care.
\end{convverbatim}
\end{tcolorbox}
\caption{Example discharge conversation generated by Qwen2.5-32B.}
\label{fig:example-qwen32b}
\end{figure*}

\begin{figure*}[t]
\centering

% --- 统一盒子风格 ---
\tcbset{
  enhanced,
  fonttitle=\bfseries,
  colframe=black!50,
  colback=white,
  sharp corners,
  boxrule=0.4pt,
  left=0.5mm,right=0.5mm,top=0.5mm,bottom=0.5mm,
  title style={left color=black!85!white, right color=black!70!white, colupper=white},
  attach boxed title to top left={yshift=-0.5mm, xshift=0.5mm},
  boxed title style={sharp corners, boxrule=0pt, colframe=black},
  before upper={\tiny}, % 全局字体缩小
}

\setlength{\parskip}{0pt} % 段落间距
\setlist[itemize]{leftmargin=3mm,itemsep=0.5pt,parsep=0pt,topsep=0.5pt} % 列表间距

\begin{minipage}{\textwidth}
\tiny
\begin{tcolorbox}[title={Patient \#1}]
\textbf{Discharge Diagnosis}\\
PRIMARY: Post-operative recovery from brain tumor removal in the fourth ventricle.\\
SECONDARY: ---\\

\textbf{Discharge Medications}
\begin{itemize}[leftmargin=4mm,itemsep=1pt]
  \item Bisacodyl 5 mg (EC): 2 tablets PO daily PRN constipation.
  \item Senna 8.6 mg: BID PRN constipation as prescribed.
  \item Docusate Sodium 100 mg: 1 tablet PO BID while on pain medications.
  \item Nifedipine SR 30 mg: 1 tablet PO HS.
  \item Oxycodone–Acetaminophen: q4--6h PRN pain as prescribed.
  \item Camphor–Menthol 0.5\%/0.5\% lotion: Apply TID PRN.
  \item Sodium Chloride 0.65\% aerosol spray: PRN up to QID.
  \item Saline sensitive eye drops: 5 times/day PRN.
\end{itemize}

\textbf{Discharge Condition}\\
Mental Status: Clear and coherent. \quad
Level of Consciousness: Alert and interactive. \quad
Activity: Ambulatory—independent.\\

\textbf{Discharge Instructions}\\
Gentle rehabilitative exercises per neurologist; maintain balanced diet and hydration; avoid strenuous activity until cleared. Seek care for severe headache, confusion, difficulty speaking, or new neurologic deficits.\\

\textbf{Follow-up Instructions}\\
Follow up with neurologist for rehabilitation and ongoing care.
\end{tcolorbox}
\end{minipage}\hfill

\begin{minipage}{\textwidth}
\tiny
\begin{tcolorbox}[title={Patient \#2}]
\tiny
\textbf{Discharge Diagnosis}\\
PRIMARY: Lower GI bleed (likely liver-disease related); Hepatic encephalopathy.\\
SECONDARY: Liver disease with portal hypertension and prior variceal bleeding.\\

\textbf{Discharge Medications}
\begin{itemize}[leftmargin=4mm,itemsep=1pt]
  \item Rifaximin 550 mg: 1 tablet PO BID.
  \item Pantoprazole 40 mg: 1 tablet PO q24h.
  \item Thiamine 100 mg: 1 tablet PO daily.
  \item Folic Acid 1 mg: 1 tablet PO daily.
  \item Calcium Carbonate 500 mg: 1 tablet PO BID (chew).
  \item Multivitamin: 1 tablet PO daily with food.
  \item Lactulose 10 g/15 mL: 30 mL PO q4h.
  \item Furosemide 20 mg: 3 tablets PO AM and 2 tablets PO PM.
  \item Spironolactone 50 mg: 3 tablets PO BID.
  \item Lidocaine HCl 2\% solution: 2 mL oral swish PRN discomfort.
  \item Acetaminophen 500 mg: 1 tablet PO q6h PRN pain (max 4 tablets/day).
  \item Ferrous Sulfate 300 mg (60 mg iron): 1 tablet PO daily on empty stomach.
\end{itemize}

\textbf{Discharge Condition}\\
Mental Status: Clear; understands instructions. \quad
Level of Consciousness: Alert and interactive. \quad
Activity: Light activity as tolerated.\\

\textbf{Discharge Instructions}\\
Strict alcohol avoidance; follow medication plan for encephalopathy and fluid retention; low-sodium diet; light activity. Seek care if symptoms worsen.\\

\textbf{Follow-up Instructions}\\
Follow up with PCP as advised; complete recommended labs; close monitoring of progress.
\end{tcolorbox}
\end{minipage}

% ---------- Row 3 (full width) ----------
\begin{minipage}{\textwidth}
\tiny
\begin{tcolorbox}[title={Patient \#3}]
\textbf{Discharge Diagnosis}\\
PRIMARY: Liver Abscess; Gram-negative bacteremia.\\
SECONDARY: ---\\

\textbf{Discharge Medications}
\begin{itemize}[leftmargin=4mm,itemsep=1pt]
  \item Metronidazole 500 mg: 1 tablet PO TID for 4 weeks (Klebsiella infection).
  \item Aspirin 81 mg: 1 tablet PO daily.
  \item Dorzolamide 2\% eye drops: 1 drop OU BID.
  \item Hydrocodone–Acetaminophen: PRN pain.
  \item Latanoprost 0.005\% eye drops: 1 drop OU nightly.
  \item Levothyroxine Sodium 25 mcg: 1 tablet PO daily.
  \item Metformin 1000 mg: 1 tablet PO BID.
  \item Metoprolol Succinate XL 12.5 mg: 1 tablet PO nightly.
  \item Nitroglycerin SL 0.3 mg: PRN chest pain.
  \item Ondansetron: PRN nausea.
  \item Pantoprazole 40 mg: 1 tablet PO daily.
  \item Timolol Maleate 0.25\% eye drops: 1 drop OU BID.
\end{itemize}

\textbf{Discharge Condition}\\
Mental Status: Clear and coherent. \quad
Level of Consciousness: Alert and interactive. \quad
Activity: Ambulatory—independent; avoid alcohol.\\

\textbf{Discharge Instructions}\\
Admitted with liver abscess and gram-negative bacteremia; treated with antibiotics and abscess drainage. Continue medication regimen, monitor for signs of infection, and keep follow-up appointments.\\

\textbf{Follow-up Instructions}\\
Follow up with PCP and our office for evaluation of underlying liver cancer and other conditions; reassess abscess and remove pigtail catheter once infection is fully cleared.
\end{tcolorbox}
\end{minipage}

\caption{Three example discharge summaries generated by Qwen2.5-72B.}
\label{fig:appendix-qwen72b-5patients}
\end{figure*}

\end{document}